\pdfoutput=1
\documentclass[letterpaper]{article} 
\usepackage{aaai23}  
\usepackage{times}  
\usepackage{helvet}  
\usepackage{courier}  
\usepackage[hyphens]{url}  
\usepackage{graphicx} 
\usepackage{natbib}  
\usepackage{caption} 
\usepackage{algorithm}
\usepackage{algorithmic}

\usepackage{tikz}
\usepackage{comment}
\usepackage{amsmath,amssymb} 
\usepackage{color}
\usepackage{makecell}
\usepackage{multirow}
\usepackage{setspace}   
\usepackage{bbding}
\usepackage{subfigure}
\usepackage{caption}
\usepackage{amsfonts}
\newcommand{\tickYes}{\ding{51}}
\usepackage{pifont}
\newcommand{\tickNo}{\ding{55}}
\usepackage{appendix}

\usepackage{newfloat}
\usepackage{listings}
\DeclareCaptionStyle{ruled}{labelfont=normalfont,labelsep=colon,strut=off} 
\floatstyle{ruled}
\newfloat{listing}{tb}{lst}{}
\floatname{listing}{Listing}
\title{FreeEnricher: Enriching Face Landmarks without Additional Cost}
\author{
    Yangyu Huang, Xi Chen, Jongyoo Kim\thanks{Corresponding author}, Hao Yang, Chong Li, Jiaolong Yang, Dong Chen\\
}
\affiliations{
    Microsoft Research Asia\\

    \{yanghuan, xichen6, jongk, haya, chol, jiaoyan, doch\}@microsoft.com,
    
}

\usepackage{bibentry}

\begin{document}

\maketitle

\begin{abstract}
Recent years have witnessed significant growth of face alignment.
Though dense facial landmark is highly demanded in various scenarios, e.g., cosmetic medicine and facial beautification, most works only consider sparse face alignment.
To address this problem, we present a framework that can enrich landmark density by existing sparse landmark datasets, e.g., 300W with 68 points and WFLW with 98 points.
Firstly, we observe that the local patches along each semantic contour are highly similar in appearance.
Then, we propose a weakly-supervised idea of learning the refinement ability on original sparse landmarks and adapting this ability to enriched dense landmarks.
Meanwhile, several operators are devised and organized together to implement the idea.
Finally, the trained model is applied as a plug-and-play module to the existing face alignment networks.
To evaluate our method, we manually label the dense landmarks on 300W testset.
Our method yields state-of-the-art accuracy not only in newly-constructed dense 300W testset but also in the original sparse 300W and WFLW testsets without additional cost.
\end{abstract}

\section{Introduction}
Face alignment provides semantically consistent facial landmarks in given face photos, which plays a critical role in several face-related vision tasks, such as face recognition \cite{masi2018deep}, face parsing \cite{luo2012hierarchical}, and face reconstruction \cite{jiang2005efficient}.
Due to its importance, great efforts are devoted to the development of algorithms as well as the construction of datasets.
Recently, remarkable alignment accuracy could be achieved through deep neural networks \cite{wu2018look,wang2019adaptive,kumar2020luvli}. 
To successfully train such deep models, a large number of face alignment datasets have been published, e.g., COFW \cite{burgos2013robust}, 300W \cite{sagonas2013300} and WFLW \cite{wu2018look}. 

Despite of enormous growth in face alignment, there have been very few researches handling dense facial landmarks.
In regard to real-world applications, dense landmarks are highly demanded in a lot of scenarios, for instance, measuring facial metrics in a medical application and delicate face editing and reenactment.
The fundamental bottleneck is the substantial cost of labeling dense facial landmarks.
To tackle this problem, we propose a novel framework for enriching landmarks\footnote{We denote \emph{enriching landmarks} by increasing landmark points along the landmark contours.} to benefit both academic and industrial scenarios.

In the literature, a 3D face model is introduced to handle dense face alignment, where the 3D model is fit to images to map each pixel to the 3D surface of the face template \cite{zhu2016face,liu2017dense,liu2016joint,cootes1995active}. 
By defining the landmark positions on the 3D face template, they could provide dense landmarks. However, these approaches suffer from the limited capacity of the 3D face model and the 2D to 3D ambiguity of the semantic landmark definition, which results in inaccurate dense landmarks and inconsistency with 2D features.

\begin{figure}[t]
\small
	\centering
	\subfigure{
		\begin{minipage}[t]{0.32\linewidth}
    	\includegraphics[width=1\linewidth]{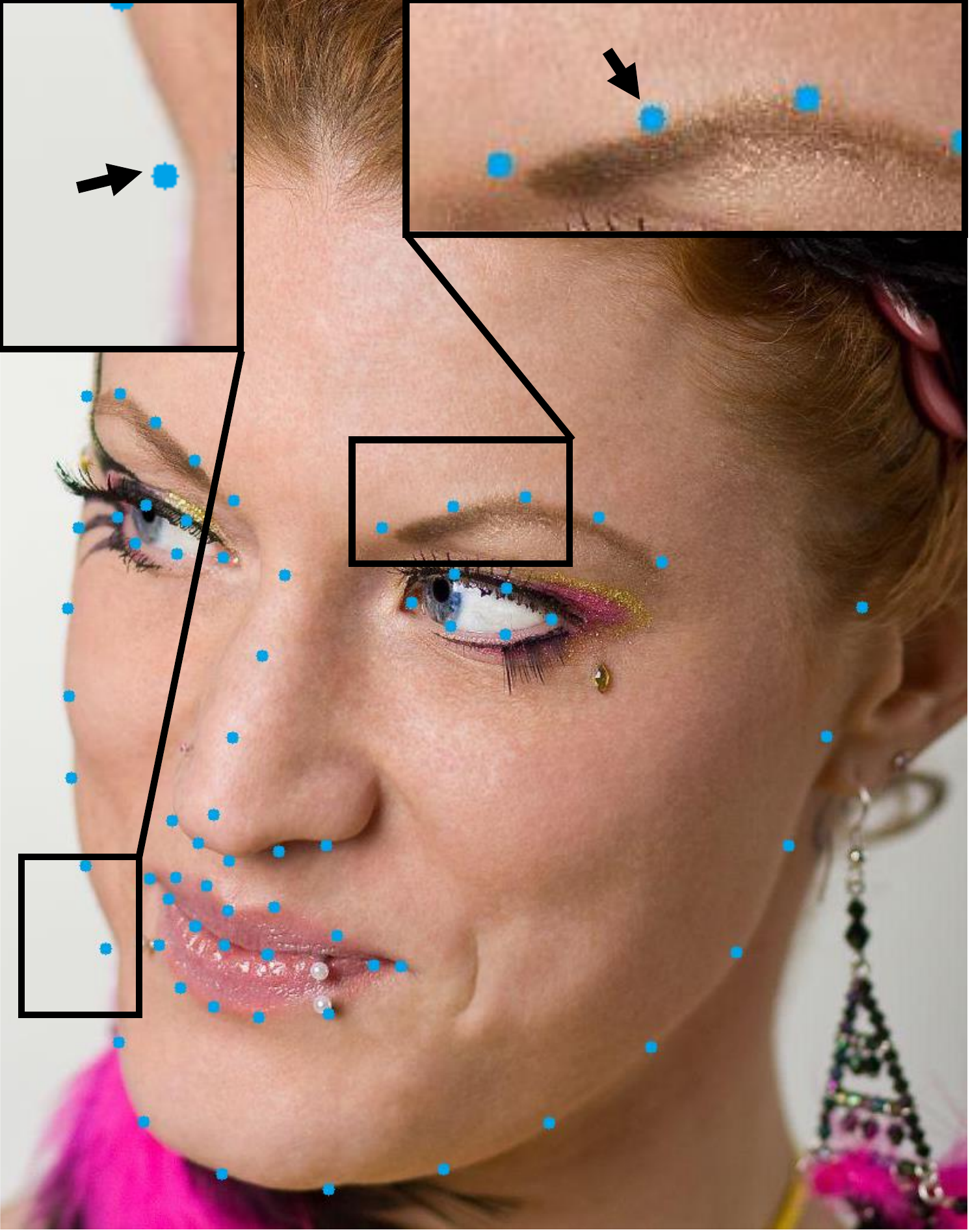}
            \captionsetup{font=scriptsize}
		\caption*{(a) ADNet (Baseline)}
		\label{figure:sparse2dense_1}
		\end{minipage}%
	}%
	\subfigure{
		\begin{minipage}[t]{0.32\linewidth}
		\includegraphics[width=1\linewidth]{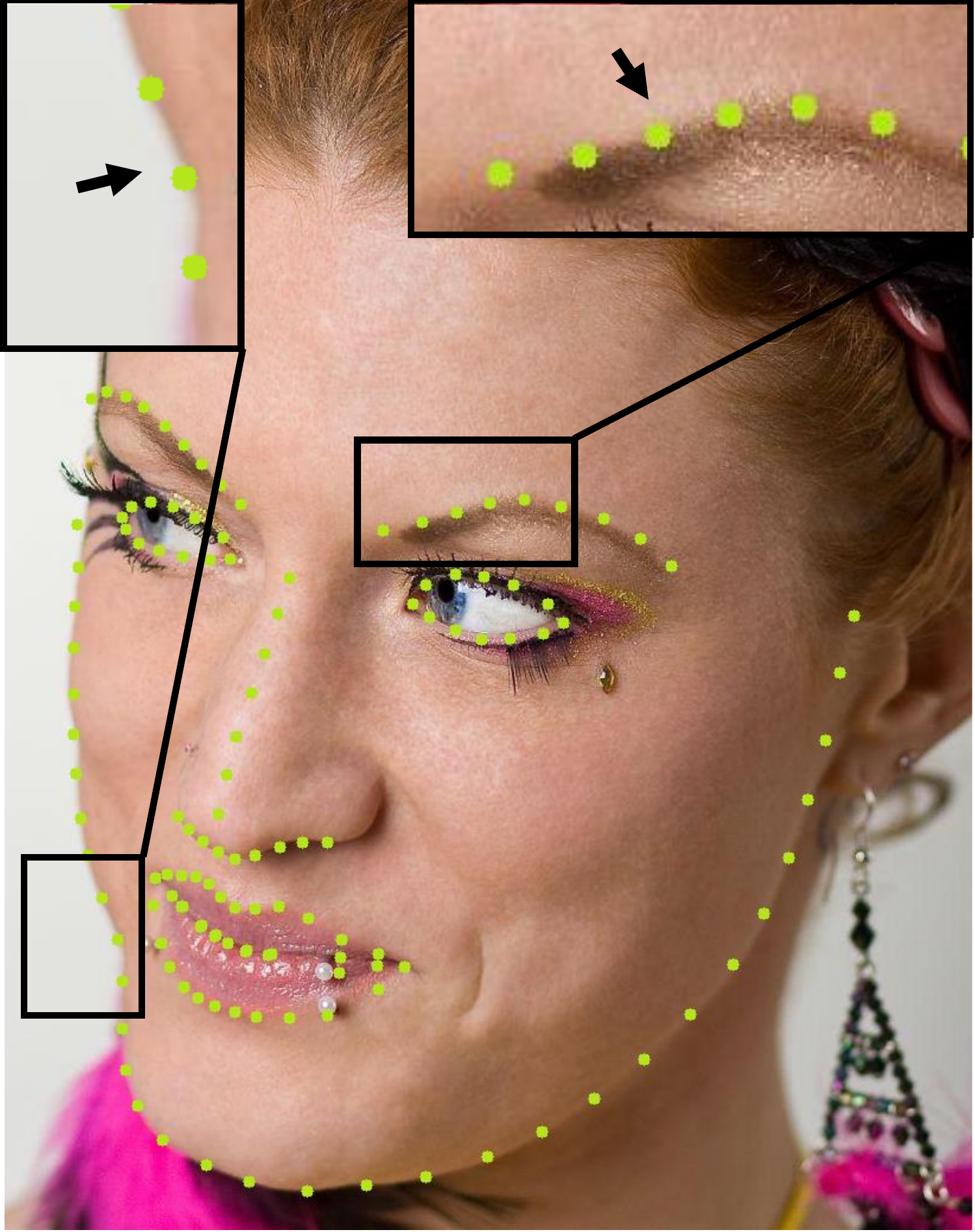}
            \captionsetup{font=scriptsize}
		\caption*{(b) ADNet-FE2 (Ours)}
		\label{figure:sparse2dense_2}
		\end{minipage}%
	}%
	\subfigure{
		\begin{minipage}[t]{0.32\linewidth}
		\includegraphics[width=1\linewidth]{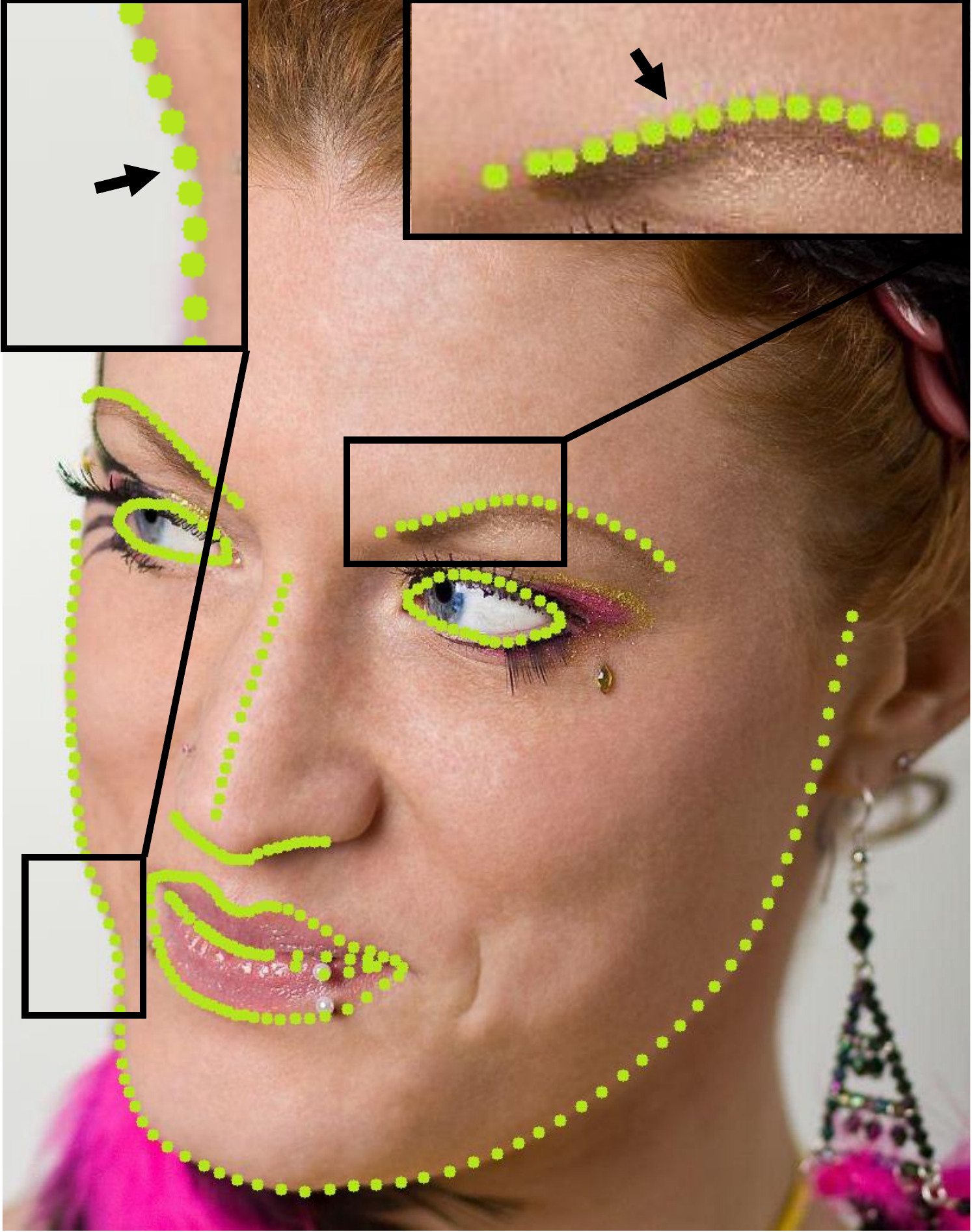}
            \captionsetup{font=scriptsize}
		\caption*{(c) ADNet-FE5 (Ours)}
		\label{figure:sparse2dense_3}
		\end{minipage}
	}%
	\centering
  \caption{
  Comparison of baseline method and our FreeEnricher method. 
  (a) is the detected sparse landmarks by baseline ADNet~\cite{huang2021adnet};
  (b) and (c) are the enriched dense landmarks by our FreeEnricher with ADNet. 
  The definitions of ADNet-FE2 and ADNet-FE5 could refer to Table~\ref{table:network_naming}. Compared to the baseline method, FreeEnricher could enrich landmarks to specific densities with higher accuracy.
  Meanwhile, FreeEnricher merely relies on the original 68-points 300W training dataset without both additional annotation and extra inference time.
}
\label{figure:sparse2dense}
\end{figure}

To mitigate these issues, we propose a weakly-supervised approach, FreeEnricher, to freely enrich the pre-annotated landmarks without any additional annotation tasks. 
Rather than relying on the 3D face model, we focus on reliable 2D features and informative landmark contours.
Specifically, we firstly observe that the local image patches are highly similar in appearance along each semantic facial component, e.g., eye, lip, and face contour.
Activated by it, we devise an idea of learning refinement ability on original sparse landmarks and transferring this ability to initialized dense landmarks.
Then a unified framework is designed to implement the idea by organizing several operators together.
Finally, FreeEnricher is plugged into the existing face alignment networks, e.g. state-of-the-art ADNet.
The new network can generate dense facial landmarks while yielding state-of-the-art accuracy.
Fig.~\ref{figure:sparse2dense} compares the results of ADNet~\cite{huang2021adnet} and our ADNet-FE2/5. While ADNet can only predict the sparse landmarks as shown in (a), ADNet-FE2 and ADNet-FE5 infer highly dense and more accurate points as shown in (b) and (c).

Our contributions can be summarized as follows:
\begin{itemize}
    \item We propose a novel method, FreeEnricher, which should be the first method that can freely enrich the facial landmark density without additional cost with respect to both manual annotation and inference time.
    \item We devise a weakly-supervised framework of FreeEnricher to transfer the refinement ability from original sparse landmarks to enriched dense landmarks. Based on it, FreeEnricher can enrich landmarks to arbitrary density and be plug-and-play to existing alignment networks.
    \item We release the enriched 300W with both preprocessed training set and annotated testing set, which would leverage the future work in the face alignment field.
    \item Our network, plugging FreeEnricher to ADNet, achieves not only accurate dense landmarks in the newly-constructed \textbf{enriched} 300W but also the state-of-the-art accuracy in the \textbf{original} 300W and WFLW testing sets.
\end{itemize}


\section{Related Work}
\label{sec:related_work} 
Face alignment has been studied widely and produced fruitful outputs. Early methods such as AAMs \cite{cootes2001active,saragih2007nonlinear,matthews2004active} and ASMs \cite{cootes1992active,cootes1995active,milborrow2008locating} are based on a generative model.
Later, the direct regression method became popular due to its simplicity and high accuracy, which can be generally categorized into coordinate and heatmap regression approaches.

\vspace{5pt}
\noindent\textbf{Coordinate Regression.}
In regression-based methods \cite{sun2013deep,toshev2014deeppose,trigeorgis2016mnemonic,lv2017deep,zhou2013extensive,zhang2014coarse}, landmark coordinates are directly regressed by the CNN network. 
To improve the localization ability of landmarks, a sort of methods \cite{zhang2014coarse,trigeorgis2016mnemonic,sun2013deep} apply coarse-to-fine strategy to refine the position.
Additionally, Smooth L1 \cite{girshick2015fast} and Wing loss \cite{feng2018wing} optimize the loss function by reducing the weight of outlier. 
Recently, methods of extra information prediction, such as uncertainties \cite{kumar2020luvli,gundavarapu2019structured}, push regression-based methods to a new climax.

\vspace{5pt}
\noindent\textbf{Heatmap Regression.}
To take full advantage of CNN networks, 2D heatmaps are utilized as intermediate outputs, then coordinates are inferred using non-maximum suppression or soft-argmax \cite{bulat2016convolutional,deng2019joint,dong2018style,newell2016stacked,wei2016convolutional}.
Several studies propose novel CNN architectures such as HRNet \cite{wang2020deep}, UNet \cite{ronneberger2015u}, and stacked HG \cite{newell2016stacked}. On the other hand, some focus on enhancing loss function. For example, Awing \cite{wang2019adaptive} addresses the sample imbalance problem between foreground and background in heatmaps.
Beyond the point-based heatmap, several studies leverage edges to enhance the accuracy of contours \cite{wu2018look,huang2020propagationnet,huang2021adnet}.

\vspace{5pt}
\noindent\textbf{Data Synthesis.}
For generating plenty of facial landmark data and annotations in model training, 
\cite{robinson2019laplace,qian2019aggregation} utilize GAN, and
\cite{huang2020ace,browatzki20203fabrec} leverage 3D reconstruction.
Both of them turned out to be effective in the lack of data.
Whereas, most of them depend on auxiliaries, such as data in other domains and 3D models.

\vspace{5pt}
\noindent\textbf{Dataset Construction.}
A number of face alignment datasets are published with diverse scenarios and various annotation definitions, but they rarely contain denser landmark annotations than 100,
such as 21 landmarks in AFLW \cite{koestinger2011annotated}, 29 landmarks in COFW \cite{burgos2013robust}, 68 landmarks in 300W \cite{sagonas2013300} and 98 landmarks in WFLW \cite{wu2018look}, due to costly annotation. 

\section{Method}
\label{sec:method}
We propose a weakly-supervised framework, called FreeEnricher, which can enrich the facial landmarks by given sparse landmarks.
Motivated by our observation that the appearance of local patches on each facial contour are similar,
the main idea is proposed that learning the refinement ability of original sparse landmarks on local patches and transfering this ability to enriched dense landmarks.
Finally, the FreeEnricher can be plug-and-play to existing face alignment networks and enrich landmarks to arbitrary density without additional cost, e.g. denser landmark annotation, and more inference time.

\subsection{FreeEnricher Framework}
\begin{figure*}[h]
\vspace{-10px}
\begin{center}
\includegraphics[width=6.8in]{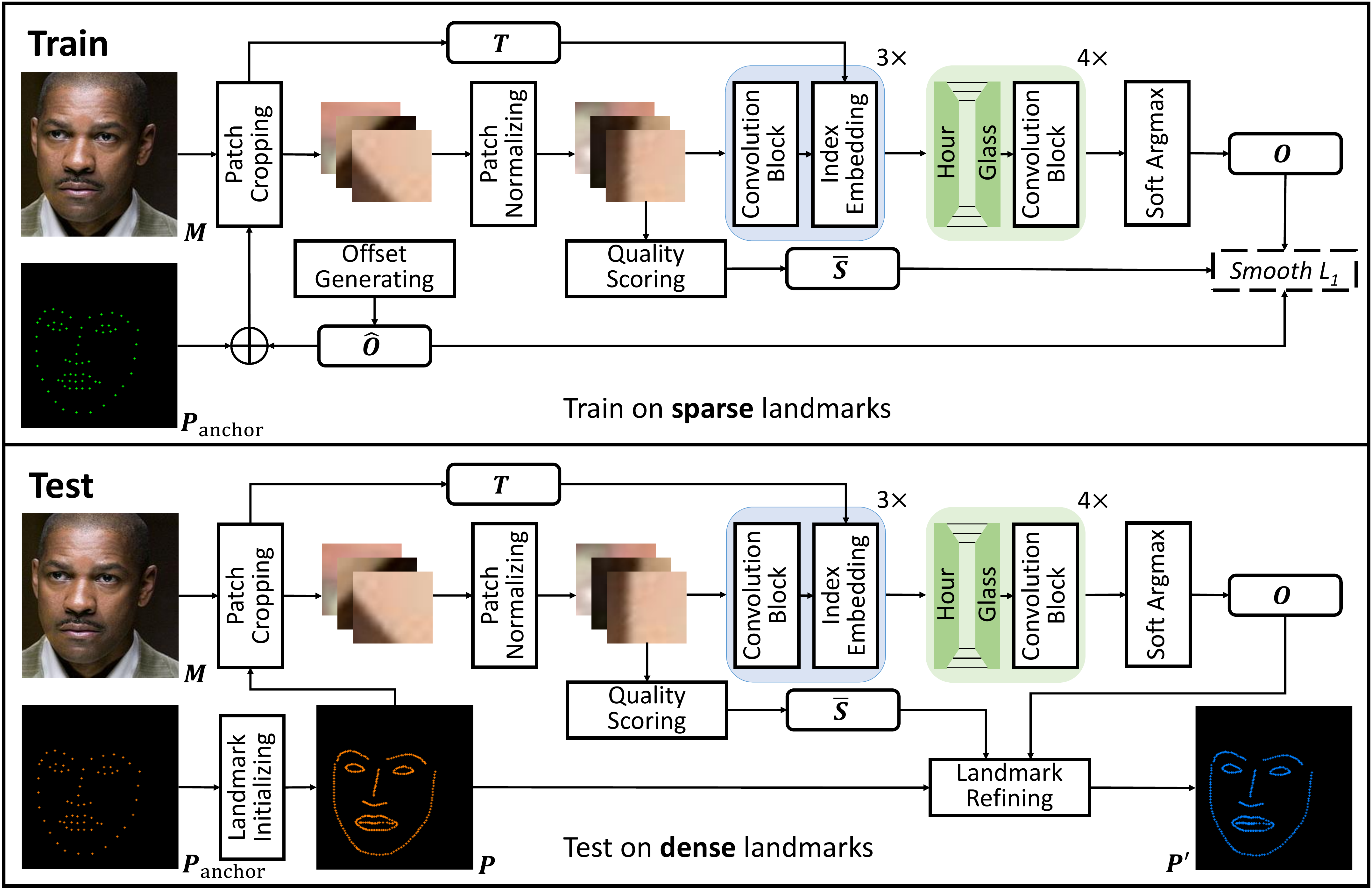}
\end{center}
   \caption{
   The framework of FreeEnricher.
   The framework employs the proposed weakly-supervised idea to enrich landmarks that training on the original sparse landmarks and testing on the enriched dense landmarks.
   In training stage, as shown in the upper part of the figure, a random offset is supervised on each normalized patch per sparse landmark and applies the normalized quality score as weight in loss function.
   In testing stage, as shown in the lower part of the figure, the enriched landmarks are initialized firstly, then refined by the trained model for each of them.
   Briefly explaining symbols, $M$ is the face image, $P_{anchor}$ is the original landmarks, $P$ is the initialized enriched landmarks, $P'$ is the final enriched landmarks, $\hat{O}$ is the randomly generated offset, $O$ is the regressed offset, $T$ is the soft index, and $\bar{S}$ is the normalized quality score.
}
\label{figure:FreeEnricher_Arc}
\vspace{-10px}
\end{figure*}

To construct the unified framework, several operators are devised and organized together in Fig.~\ref{figure:FreeEnricher_Arc}.
They are \emph{landmark initializing}, \emph{offset generating}, \emph{patch cropping}, \emph{patch normalizing}, \emph{quality scoring}, and \emph{index embedding}.
A detailed description of each operator could be referred to below.

\vspace{5pt}
\noindent\textbf{\emph{Landmark Initializing}}
initialize the rough enriched landmarks from sparse landmarks by averagely interpolating points from each facial contour, such as eye, mouth, and face contour.
Hence, straight-line fitting and b-spline fitting \cite{knott2000interpolating} could be selectively applied to fit the contours, and we call each fitted contour $C$.
Given the enriching density $D$ of \emph{landmark initializing} and the original landmarks $P_{anchor}$, the rough enriched landmarks $P$ of each contour are indicated as

\vspace{-5pt}
\begin{equation}
p_{i,j} = 
\begin{cases}
\ C(u_{i,j}),& j\in \{1,2,..., D{-}1\} \\
\ C(u_i),& j=0 \\
\end{cases}
\label{equation:landmarks_enrich}
\end{equation}
\vspace{-5pt}

where $i$ is the index of \emph{anchor point}, $j$ is the index of \emph{interpolated point} between two adjacent \emph{anchor points} and $u$ controls the relative position of enriched point in curve.
We denote the original landmark as \emph{anchor point}, where $j=0$, and the newly sampled point as \emph{interpolated point}, where $j>0$.
Then the $u$ of \emph{interpolated point} is derived by

\vspace{-5pt}
\begin{equation}
u_{i,j} = \frac{D-j}{D} \cdot u_{i} + \frac{j}{D} \cdot u_{i+1}
\label{equation:upsampler}
\end{equation}
\vspace{-5pt}

where $u_i$ and $u_{i+1}$ present two adjacent \emph{anchor points}.
Since the curve is derived by only considering the existing sparse landmarks in geometry, the newly generated points are still not consistent with the image texture.

\vspace{5pt}
\noindent\textbf{\emph{Offset Generating}}
simulates the small position errors of \emph{interpolated points} in normal by applying a random 2D offset for each \emph{anchor point} in normal direction during training stage.
We denote the generated offset by $\hat{O}$, which is the ground truth.
Given the local patch size $s_{patch}$, the offset follows the uniform distribution $U(-s_{patch}/8, +s_{patch}/8)$.

\vspace{5pt}
\noindent\textbf{\emph{Patch Cropping}}
crops local image patch from the face image at the center of input point by the size of $s_{patch}{\times}s_{patch}$, where the input point denotes distorted anchor points $P_{anchor}+\hat{O}$ in training stage and initialized enriched landmarks $P$ in testing stage.
The relative scale of cropped patch is controlled by the patch-face ratio which represents the ratio of patch size to aligned face size.

\begin{figure}[h]
\begin{center}
\includegraphics[width=3.3in]{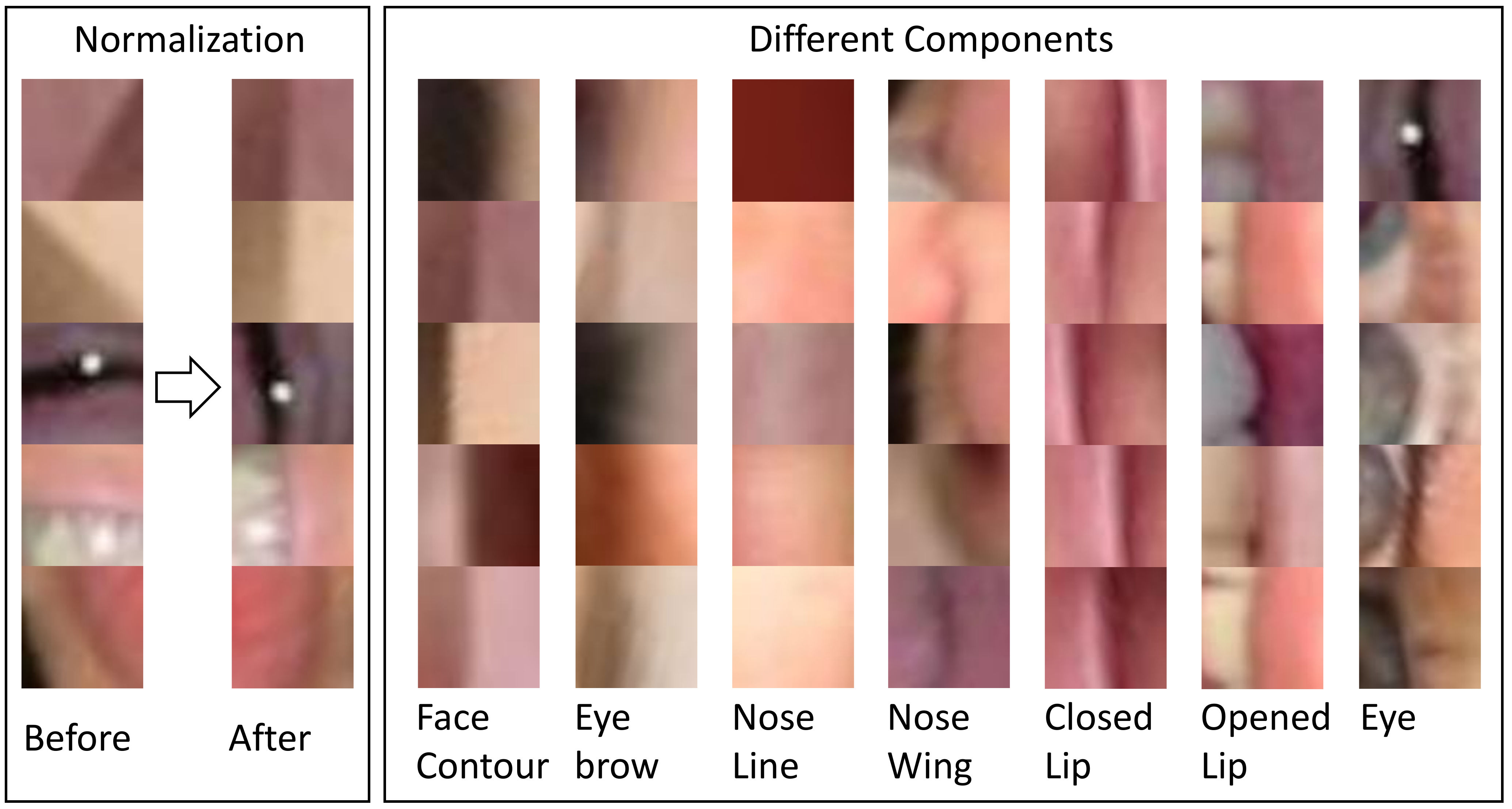}
\end{center}
   \caption{Patch samples that $64 \times 64$ region is cropped from $1024 \times 1024$ aligned face image (patch-face ratio is 1/16). The left 2 columns show the patches before/after normalization. The right 7 columns show the typical patches from different facial components.
   }
\label{figure:patch_samples}
\end{figure}

\vspace{5pt}
\noindent\textbf{\emph{Patch Normalizing}}
rotates each patch so that its $x$-axis (horizontal direction) is aligned with the normal of the corresponding landmark along the contour line, which is illustrated in the left 2 columns of Fig.~\ref{figure:patch_samples}. 
The operator is based on the assumption that the precision improvement along the normal direction is more effective and reasonable than the tangent direction \cite{huang2021adnet}.
Furthermore, the freedom of the regression target is reduced from 2D to 1D for no ambiguity and better convergence.

\vspace{5pt}
\noindent\textbf{\emph{Quality Scoring}}
is designed to handle blurry boundaries, occluded regions, and contour corners.
The goal of this operator is to provide the reliability of each patch, which is used to weight the losses in training and provide confidence in testing. The intermediate value $V$ of the score is derived by

\vspace{-5pt}
\begin{equation}
V(k) = \frac{\sigma(\sum\limits_{y=1}^{h}patch_k(x,y))}{\sigma(\sum\limits_{x=1}^{w}patch_k(x,y))}
\label{equation:score_origin_1}
\end{equation}
\vspace{-5pt}

where $w$ and $h$ are the width and height of the local image patch, $patch_k(x,y)$ is the gray value of the specific pixel in $k$th patch and $\sigma(\cdot)$ is the standard deviation.
Equation~\ref{equation:score_origin_1} is maximized when the variance along the horizontal direction is high, and the variance along the vertical direction is low, which aims to find a clear vertical boundary in the aligned image patch, in other words, those patches are suitable to regress the accurate rectification of the landmark because of the similar appearance and clear edge feature.
The raw score $S$ of the patch is then obtained by
\begin{equation}
S(k) = 
\begin{cases}
\ V(k) - 1,& V(k) >= 1 \\
\ 1 - \frac{1}{V(k)},& otherwise \\
\end{cases}
\label{equation:score_origin_2}
\end{equation}

\begin{figure}[h]
\begin{center}
\includegraphics[width=3.3in]{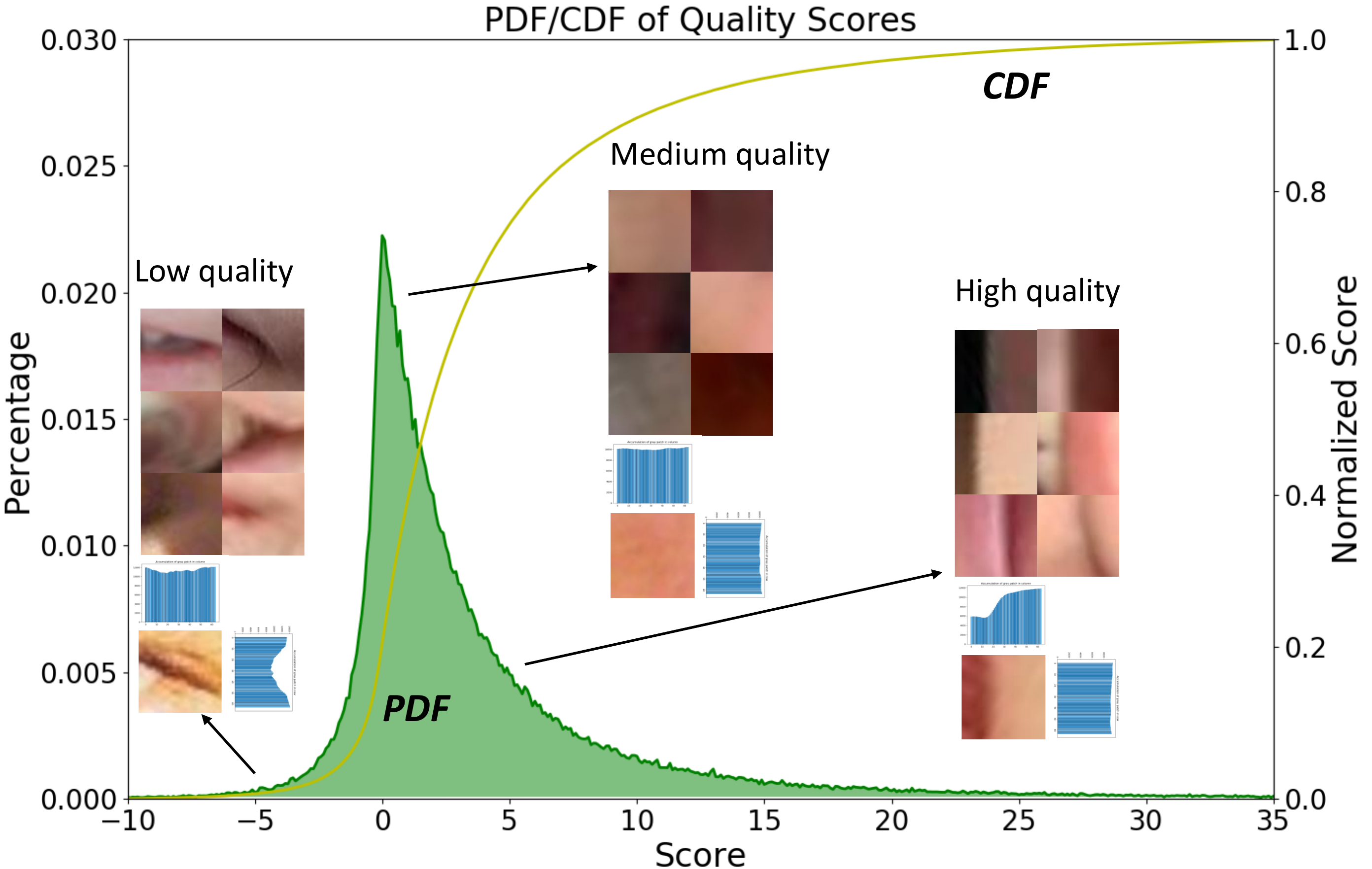}
\end{center}
   \caption{PDF/CDF of patch quality score in 300W training dataset. The green distribution stands for the probability density function (PDF) of scores whose y-axis is at left, and the yellow curve indicates the cumulative distribution function (CDF) of scores whose y-axis is at right. The CDF is also the mapping function from score to normalized score. The patches are divided into low-quality, medium-quality, and high-quality category. Their corresponding typical patch samples are exhibited by category. Meanwhile, the distributions of pixel values in column and in row are also presented for a better understanding of the quality scoring.}
\label{figure:score_distribution}
\end{figure}

To normalize the raw score within 0 and 1, we firstly build a cumulative distribution function (CDF) using the whole patch scores in the training set, then map each raw score to the final normalized score $\bar{S}$ by the cumulated probability in Equation~\ref{equation:score_norm}. 
\begin{equation}
\bar{S}(k) = \frac{1}{N} \cdot \sum\limits_{n=1}^{N} 
\begin{cases}
\ 1 ,& S(n) < S(k) \\
\ 0 ,& otherwise \\
\end{cases}
\label{equation:score_norm}
\end{equation}

where $N$ is the total number of patches in the training dataset.
The distribution of the quality scores and their patch samples are shown in Fig.~\ref{figure:score_distribution}, which verifies the rationality in visual.
In general, the aligned patch with a high-quality score has a clear vertical boundary, with a medium-quality score has a blurry boundary or occlusion, and with a low-quality score has a contour corner.

\vspace{5pt}
\noindent\textbf{\emph{Index Embedding}} provides the position code of each patch similarity to transformer network \cite{vaswani2017attention}.
Specifically, we assign a soft index (position) to each patch (word) on the face (sentence).
The right 7 columns of Fig.~\ref{figure:patch_samples} show that the patches in the same component are similar, but there still exists a marginal appearance gap across components.
Meanwhile, we can not put the patches of each landmark to the model in order, because of the misaligned landmark density between training and testing.
For the above reasons, the \emph{index embedding} in our framework is necessary to further improve the generalization of the model.

Assuming that the feature maps $\bold{F}_{in}$ from the previous CNN layer have $m$ channels, then $\{\lfloor t \rfloor, n + \lfloor t \rfloor, 2n + \lfloor t \rfloor, ..., m - n + \lfloor t \rfloor\}$ indices are selected as $\bold{U}$ to filter channels from input feature maps, where $n$ is the number of \emph{anchor points}. Given the soft index $t$, the output feature maps $\bold{F}_{out}$ of \emph{index embedding} are presented as
\begin{equation}
\bold{F}_{out} = (1 + \lfloor t \rfloor - t)\cdot \bold{F}_{in}(\bold{U}) + (t - \lfloor t \rfloor) \cdot \bold{F}_{in}(\bold{U} + 1)
\label{equation:index_embed}
\end{equation}

To bridge the different landmark density between the training and testing stages of FreeEnricher, the soft index $t$ of $p_{i,j}$ could be calculated by
\begin{equation}
T(i,j) = i + \frac{j}{D}
\label{equation:pos}
\end{equation}
where $p_{i,j}$ is the enriched landmark. 
Specifically, $p_{i,j}$ is the anchor point, when $j = 0$, otherwise, the interpolated point.

\vspace{5pt}
\noindent\textbf{\emph{Landmark Refining}}
refines the initialized enriched landmarks $P$ by adding the regressed offset $O$ with normalized quality score $\bar{S}$ as confidence (weight), which denotes as
\begin{equation}
\begin{aligned}
P' = P + \bar{S} \times O
\end{aligned}
\label{equation:landmarks_refine}
\end{equation}
where $P'$ is the final refined enriched landmarks by FreeEnricher in inference.

\vspace{5pt}
\noindent\textbf{Training process of FreeEnricher}
is presented in the upper part of Fig.~\ref{figure:FreeEnricher_Arc}, by inputting the face image $M$ and original landmarks $P_{anchor}$, we firstly apply the \emph{offset generating} to the original landmarks $P_{anchor}$ to generate the ground truth of offsets $\hat{O}$,
which would distort the original landmarks $P_{anchor}$,
then feed the face image $M$ with distorted original landmarks $P_{anchor}+\hat{O}$ to model (start from \emph{patch cropping} and end in \emph{soft argmax}) to get the regressed offsets $O$ and normalized quality scores $\bar{S}$, finally put them ($\bar{S}$, $O$, $\hat{O}$) to the loss function of FreeEnricher $\mathcal{L}$, which is defined as

\begin{equation}
\mathcal{L} = \frac{1}{B} \cdot \sum\limits_{i=1}^{B} \bar{S}(i) \cdot Smooth\ L_1(O(i), \hat{O}(i))
\label{equation:patch_loss}
\end{equation}

where $B$ is the batch size, $O$ is the regressed offset, $\hat{O}$ is the generated offset, and $\bar{S}$ is the normalized quality score.


\vspace{5pt}
\noindent\textbf{Testing process of FreeEnricher}
is demonstrated in the lower part of Fig.~\ref{figure:FreeEnricher_Arc}, by inputting the face image $M$ and original landmarks $P_{anchor}$, we firstly employ the \emph{landmark initializing} on the original landmarks $P_{anchor}$ to generate initialized enriched landmarks $P$, then feed the face image $M$ with initialized enriched landmarks $P$ to model (start from \emph{patch cropping} and end in \emph{soft argmax}) to get the regressed offsets $O$ and normalized quality scores $\bar{S}$, finally put them ($P$, $\bar{S}$, $O$) to \emph{landmark refining} to acquire the final accurate enriched landmarks $P'$.


\subsection{FreeEnricher Network}

The model from FreeEnricher framework has the ability to enrich sparse landmarks to accurate dense landmarks.
To make full use of this ability, we apply the FreeEnricher model to the
existing face alignment networks (baseline networks) by plug-and-play mode,
then get the FreeEnricher networks (enhanced networks).
Specifically, there are two ways to plug in the model: (1) employing the model to pre-process the training dataset, which means densifying the ground truth of sparse landmarks, and (2) utilizing the model to postprocess the testing results, which denotes densifying the predicted value of sparse landmarks.
Each way can be used optionally, thus we define the following 3 modes.

\vspace{5pt}
\noindent\textbf{Plug in Test}
denotes that FreeEnricher only enriches the predicted landmarks in the testing stage.
Under this mode, the total inference time increases tremendously.

\vspace{5pt}
\noindent\textbf{Plug in Train}
indicates that FreeEnricher merely enriches the ground truth of landmarks in the training stage.
Under this mode, there is no extra computational cost in inference, and it is the default mode in our method.

\vspace{5pt}
\noindent\textbf{Plug in Train and Test} means that FreeEnricher model enriches the ground truth of landmarks in the training stage with \emph{landmark initializing} and refines the predicted landmarks in the testing stage without \emph{landmark initializing}.
Under this mode, the network always gets the highest accuracy, but it is pretty time-consuming in practical deployment.

\begin{table}[htbp]
\begin{center}
\scriptsize
    \begin{tabular}{lcccc}
    \hline
\makecell{FreeEnricher \\ Network} & \makecell{Baseline \\ Network} & \makecell{Enriching \\ Density} & \makecell{Plug in \\ Train} & \makecell{Plug in \\ Test} \\
    \hline
    \textbf{ResNet50-FE2$_{train}$} & ResNet50 & 2 & \tickYes & \tickNo \\
    \textbf{HRNet-FE3$_{test}$} & HRNet & 3 & \tickNo & \tickYes \\
    \textbf{ADNet-FE5} & ADNet & 5 & \tickYes & \tickNo \\
    \textbf{ADNet-FE10$_{train+test}$} & ADNet & 10 & \tickYes & \tickYes \\
    \hline
    \end{tabular}
\end{center}
\caption{
FreeEnricher network samples. 
}
\label{table:network_naming}
\end{table}

The name of FreeEnricher network is defined as "$BaselineNet$-FE$D_{stage}$" in a unified format, where $BaselineNet$ indicates which face alignment baseline network to be used, $D$ denotes the enriching density of \emph{landmark initializing} and $stage$ means in which stage the FreeEnricher plug, $stage$ could be omitted when its value is "train".
Several FreeEnricher network samples are presented in Table~\ref{table:network_naming} for clearer understanding.

\section{Experiment}
\label{sec:experiment}

\subsection{Implementation Details}
\label{sec:experiment_setting}


\vspace{5pt}
\noindent\textbf{FreeEnricher.} 
The FreeEnricher model is trained on four GPUs (16GB NVIDIA Tesla P100) by PyTorch~\cite{paszke2019pytorch}, where the batch size of each GPU is 68 for 300W and 98 for WFLW.
We employ Adam optimizer with the initial learning rate of $1\times10^{-3}$ and decay the learning rate by $1/10$ at the epochs of 100, 150, and 180, finally ending at 200.
Specifically,
\emph{landmark initializing} adopts b-spline interpolation method implemented in Scipy \cite{virtanen2020scipy} with order of 3 and enriching density of 5,
\emph{offset generating} randomly generates offset by the uniform distribution $U(-8, +8)$, and
\emph{patch cropping} crops $64{\times}64$ patch from aligned face image with $1024{\times}1024$ resolution (patch-face ratio of 1/16) by the center of target landmark.
The normalized patches are augmented by simulated random gray, random blur, and random occlusion.
Regarding the backbone architecture of FreeEnricher, we adopt 3-stacked \emph{index embedding} modules and 4-stacked \emph{hourglass} modules,
where each \emph{hourglass} outputs a $1 \times 64 \times 64$ feature map.
The output feature maps are then fed to $soft\ argmax$ \cite{huang2021adnet} to generate the landmark offsets. The final landmark offset is from the last \emph{hourglass}.
The following experiments employ the settings above if no further explanation.

\vspace{5pt}
\noindent\textbf{Face Alignment Network.} The FreeEnricher model enhances the existing face alignment network by "Plug in Train" mode in both landmark density and accuracy. And all of the existing face alignment networks are trained using the identical setting to the original papers.

\begin{figure}
\centering
\includegraphics[width=1\linewidth]{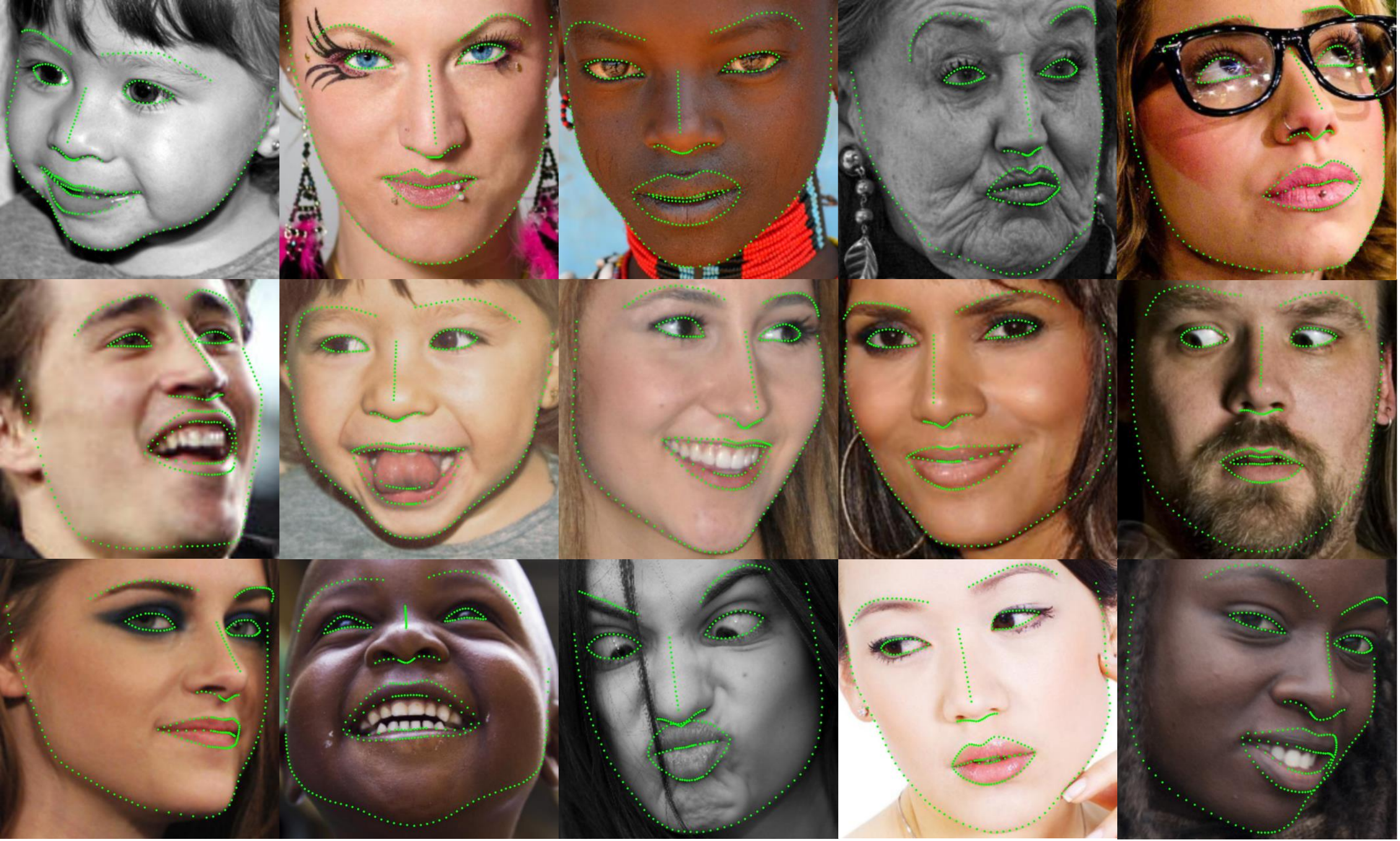}
\caption{Annotation samples of the \textbf{enriched} 300W benchmark with 320 landmarks. Those annotations are newly labeled by experts.}
\label{figure:300W_benchmark_samples}
\end{figure}

\vspace{5pt}
\noindent\textbf{Enriched 300W test set.} We newly construct an enriched 300W test dataset by manually labeling the original images, which is employed to evaluate the performance of enriched face alignment. 
To handle different densities of landmarks, we labeled the continuous curve rather than discrete points.
From the curve, firstly, we extract the anchor points by finding the points (on the curve) which are closest to the original landmark labels in 300W.
Then, between two neighboring anchor points, the new points are uniformly sampled.
For instance, from the original 68 landmarks, 320 enriched landmarks are generated with the enriching density 5.
Typical samples are shown in Fig.~\ref{figure:300W_benchmark_samples}.

\subsection{Evaluation Metrics}
\label{sec:experiment_metrics}

\noindent\textbf{Mean Error (ME)} measures the averaged $L2$ distance between predicted points and ground-truth points, which is specifically employed to evaluate the offset regression of fixed-scale patch in FreeEnricher model and defined as
\begin{equation}
\textnormal{ME}(P, \hat{P}) = \frac{1}{N_P}  \sum\limits_{i=1}^{N_P}\|p_i-\hat{p}_i\|_2
\label{equation:ME}
\end{equation}
where $P$ and $\hat{P}$, respectively, denotes the predicted values and the ground-truth of landmarks (or offsets), and $N_P$ is the number of landmarks (or offsets).

\vspace{5px}
\noindent\textbf{Normalized Mean Error of Points (NME$_{point}$ or NME)} between predicted points and ground-truth points is a widely-used standard metric to evaluate landmark accuracy for face alignment, which is defined as
\begin{equation}
\textnormal{NME}_{point}(P, \hat{P}) = \frac{\textnormal{ME}(P, \hat{P})}{d}
\label{equation:NME_point}
\end{equation}
where $d$ is the unit distance used to normalize the errors. Inter-ocular distance (distance between outer eye corners) is employed as the unit distance in the following experiments.

\vspace{5px}
\noindent\textbf{Normalized Mean Error of Edges (NME$_{edge}$)} between predicted points and ground-truth edges is proposed to evaluate the performance of the landmarks, which emphasizes the error on the normal direction and is defined as
\begin{equation}
\begin{aligned}
\textnormal{NME}_{edge}(P, \hat{P}, \hat{E}) = 
&\frac{1}{N_P} \sum\limits_{i=1}^{\hat{p_i} \notin \hat{E}\ and\ N_P}\frac{\|p_i-\hat{p}_i\|_2}{d} + \\
&\frac{1}{N_P} \sum\limits_{i=1}^{\hat{p_i} \in \hat{E}\ and\ N_P}\frac{dist(p_i, \hat{e}_i)}{d}
\end{aligned}
\label{equation:NME_edge}
\end{equation}
where $\hat{E}$ is the ground-truth of edges, $\hat{e_i}$ is the edge that the $i$th landmark belongs to, and $dist(\cdot)$ computes point-to-edge distance.

\subsection{Comparison on Enriched Landmarks}
The main advantage of applying the FreeEnricher is that we get denser landmarks than from the original dataset.
In order to demonstrate the effectiveness of FreeEnricher on enriching landmarks, we evaluate the NME of landmarks in the newly-constructed enriched 300W testing dataset.
Because there is no existing method that could freely enrich landmarks, we design a baseline method as the comparison by applying \emph{landmark initializing} operator, which enriches the landmark density of the original network.
Meanwhile, the state-of-the-art ADNet is selected as the baseline network to be enhanced for a good starting point.
Besides, other settings are identical between these 2 experiments.
The result could be referred to Table~\ref{table:300W_enrich}, where FreeEnricher significantly enhances the accuracy of enriched landmarks, especially $\sim$20\% improvement in the normal direction. 

\begin{table}[htbp]
\begin{center}
 \begin{tabular}{llccc}
\hline
Method & Network & NME$_{point}$ & NME$_{edge}$ \\
\hline
Baseline & ADNet + Line5 & 3.21 & 1.18 \\
\hline
\textbf{Ours} & \textbf{ADNet-FE5} & \textbf{3.06} & \textbf{0.98} \\
\hline
\end{tabular}
\end{center}
\caption{Comparison between FreeEnricher and baseline method on the \textbf{enriched} 300W testing set, totally 320 landmarks. “Line5” represents applying \emph{landmark initializing} to the output of network by straight-line interpolation with $D$ of 5.}
\label{table:300W_enrich}
\end{table}

\subsection{Comparison on Original Landmarks}
Beyond enriching landmarks, the FreeEnricher can benefit the baseline network on the sparse landmarks as well.
For the same reason, we still use ADNet as the baseline.
Hence, we also evaluate FreeEnricher and the other face alignment algorithms on the original 300W and WFLW datasets in Table~\ref{table:300W_WFLW_origin}.
The results show that FreeEnricher unexpectedly improves the accuracy further, and outperforms the existing methods on both 300W and WFLW.

\begin{table}[htbp]
\begin{center}
\begin{tabular}{lcc}
\hline
Method & 300W & WFLW \\
\hline
LAB \cite{wu2018look} & 3.49 & 5.27 \\
HRNet \cite{wang2020deep} & 3.34 & 4.60 \\
LUVLi \cite{kumar2020luvli} & 3.23 & 4.37 \\
ADNet \cite{huang2021adnet} & 2.93 & 4.14 \\
\hline
\textbf{ADNet-FE5 (Ours)} & \textbf{2.87} & \textbf{4.10} \\
\hline
\end{tabular}
\end{center}
\caption{$\textnormal{NME}$ comparison between our method and other state-of-the-art on the \textbf{original} 300W and WFLW testing set, 68 and 98 landmarks in total respectively. To compare the result of ADNet-FE5 network, we merely select the original landmarks from the enriched landmarks.}
\label{table:300W_WFLW_origin}
\end{table}

\subsection{Ablation Study}
To verify the efficacy of FreeEnricher in unit and the contribution of each operator, we conduct comprehensive ablation experiments. Furthermore, they also demonstrate the robustness of different face alignment networks and the applicability to various landmark densities.

\vspace{5pt}
\noindent\textbf{Unit Verification.}
To evaluate the offset refinement network independently from the
whole system on local image patches,
we conduct unit verification in Table~\ref{table:300W_WFLW_patch}.
Hereon, FreeEnricher is trained on the generated patches from the training set and tested from the testing set respectively.
Both the results on 300W and WFLW datasets demonstrate that FreeEnricher could refine the landmark location in the local patch of the face image.

\begin{table}[htbp]
\small
\begin{center}
\begin{tabular}{lcc}
\hline
Method & \makecell{300W Testset} & \makecell{WFLW Testset} \\
\hline
Random offset & 4.03 & 3.96 \\
Regressed offset & \textbf{1.92} & \textbf{1.89} \\
\hline
\end{tabular}
\end{center}
\caption{
Unit verification of FreeEnricher on 300W and WFLW datasets. The second row indicates the ME of the generated offsets by \emph{offset generating} operator. The third row indicates the ME of the regressed offsets by FreeEnricher.
}
\label{table:300W_WFLW_patch}
\end{table}

\vspace{5pt}
\noindent\textbf{Contribution of Each Operator.}
After integrating the designed operators into the framework individually, the performance increases to different degrees, which demonstrates the contribution of each operator in FreeEnricher framework.
B-spline \emph{landmark initializing} and \emph{patch normalizing} improve more significantly than \emph{quality scoring} and \emph{index embedding} on accuracy.
The best configuration is G in Table~\ref{table:FE_operators}.

\begin{table}[htbp]
\scriptsize
\begin{center}
\begin{tabular}{lccccc}
\hline
\makecell{Config.} & \makecell{Landmark \\ Initializing} & \makecell{Patch \\ Normalizing} & \makecell{Quality \\ Scoring} & \makecell{Index \\ Embedding} & NME \\
\hline
A & - & - & - & - & 3.21 \\
\hline
B & line & \tickNo & \tickNo & \tickNo & 3.18 \\
C & b-spline & \tickNo & \tickNo & \tickNo & 3.11 \\
D & line & \tickYes & \tickNo & \tickNo & 3.14 \\
E & line & \tickNo & \tickNo & \tickYes & 3.16 \\
F & line & \tickYes & \tickYes & \tickNo & 3.11 \\
\hline
\textbf{G} & b-spline & \tickYes & \tickYes & \tickYes & \textbf{3.06} \\
\hline
\end{tabular}
\end{center}
\caption{Contribution of operator on \textbf{enriched} 300W testing dataset. A indicates average upsampling original landmarks by linear interpolation without FreeEnricher regression.}
\label{table:FE_operators}
\end{table}

\vspace{5pt}
\noindent\textbf{Robustness on Existing Network.} The FreeEnricher could be plug-and-play to any face alignment network in theory.
To verify the robustness in practice, several networks, ResNet50~\cite{zhang2014facial}, HRNet~\cite{wang2020deep}, and ADNet~\cite{huang2021adnet}, are selected to conduct the comparison experiments.
Table~\ref{table:networks} demonstrate that all of them have a significant improvement on enriched 300W testing set after applying the FreeEnricher.

\begin{table}[htbp]
\centering
\begin{tabular}{lc}
\hline
Network & NME \\
\hline
ResNet50 + Line5 \cite{zhang2014facial} & 4.76 \\
\textbf{ResNet50-FE5 (Ours)} & \textbf{4.39} \\
\hline
HRNet + Line5 \cite{wang2020deep} &  3.74 \\
\textbf{HRNet-FE5 (Ours)} & \textbf{3.46} \\
\hline
ADNet + Line5 \cite{huang2021adnet} & 3.21 \\
\textbf{ADNet-FE5 (Ours)} & \textbf{3.06} \\
\hline
\end{tabular}
\caption{Robustness study of FreeEnricher on different face alignment network on \textbf{enriched} 300W testing set.}
\label{table:networks}
\end{table}

\vspace{5pt}
\noindent\textbf{Applicability on Enriching Density}
The FreeEnricher could enrich landmarks to arbitrary density in theory.
To verify the applicability in practice, various enriching densities, 1, 2, 3, 5, and 10, are selected to conduct the verification experiments.
Table~\ref{table:density_landmarks} demonstrates that all of them achieve consistently high accuracy on the enriched 300W testing dataset by the given enriching density.
Besides, the denser the landmarks are, the larger the NME becomes.

\begin{table}[htbp]
\centering
\begin{tabular}{lcc}
\hline
\makecell{Enriching Density} & \makecell{Landmarks Count} & NME \\
\hline
1 & 68 & 2.91 \\
2 & 131 & 3.01 \\
3 & 194 & 3.03 \\
5 & 320 & 3.06 \\
10 & 635 & 3.14 \\
\hline
\end{tabular}
\caption{Applicability study of FreeEnricher on different enriching density $D$ on \textbf{enriched} 300W testing dataset.}
\label{table:density_landmarks}
\end{table}

\vspace{5pt}
\noindent\textbf{Study on Plugging Stage.}
The FreeEnricher could be plug-and-play in existing networks under their training and testing stage.
The experiments in Table~\ref{table:FE_stage} illustrate that the FreeEnricher plugged in different stages have similar accuracy improvement but different computational cost in inference.
To balance both high accuracy and low computational cost in the deployment environment, FreeEnricher is merely plugged in training stage.
Then, FreeEnricher brings improvement on enriched landmarks without extra inference time and memory.

\begin{table}[htbp]
\small
\begin{center}
\begin{tabular}{lccc}
\hline
\makecell{Network} & NME & \makecell{Extra \\ FLOPs} & \makecell{Extra \\ Params} \\
\hline
ADNet-FE5$_{test}$ & 3.07 & +9.85~G & +6.78~M \\
\textbf{ADNet-FE5$_{train}$} & 3.06 & \textbf{+0~G} & \textbf{+0~M} \\
ADNet-FE5$_{train+test}$ & \textbf{3.04} & +9.85~G & +6.78~M \\
\hline
\end{tabular}
\end{center}
\caption{Study of FreeEnricher on different plugging stage on \textbf{enriched} 300W testing dataset. The subscript of the network name denotes the plugging stage of FreeEnricher.}
\label{table:FE_stage}
\end{table}

\section{Conclusion}
\label{sec:conclusion}

Enriching facial landmarks is a meaningful topic in both academic and industry.
In this paper, we present a novel idea to address this problem by using the existing datasets without additional labeling annotation and inference time costs.
Moreover, our method could enrich landmarks to arbitrary density and be plug-and-play to any existing face alignment network in theory, which is also demonstrated by various experiments.
Specifically, a weakly-supervised framework is proposed, which learns the refinement ability on original sparse landmarks of existing datasets, then this ability is applied to the initialized dense landmarks.
By our method, we not only obtain accurate enriched facial landmarks but also achieve state-of-the-art performance in the original dataset.

\bibliography{aaai23.bib}
\nobibliography{aaai23}

\clearpage
\appendix

\section{Appendix}

In order to introduce the paper more clearly and provide more information as a reference, we add supplementary material as an appendix.
Briefly speaking, the supplementary material contains three sections,
firstly, the application of facial morphometric measurement demonstrates the significance of enriching landmarks in downstream tasks,
then, the additional details of the proposed approach are also provided including landmark contour definition, landmark contour fitting, landmark normal calculation, and regression model architecture,
finally, in order to further verify the efficacy of our method on different facial components and different extreme scenarios, several additional experiments are conducted, and the visualization result on 300W and WflW test sets furtherly proves it.

\section{Application}
\label{sec:Application}

\subsection{Facial Morphometric Measures}

\begin{figure}[htp!]
\centering
\subfigure[chin angle]{
    \centering
    \includegraphics[width=0.14\textwidth]{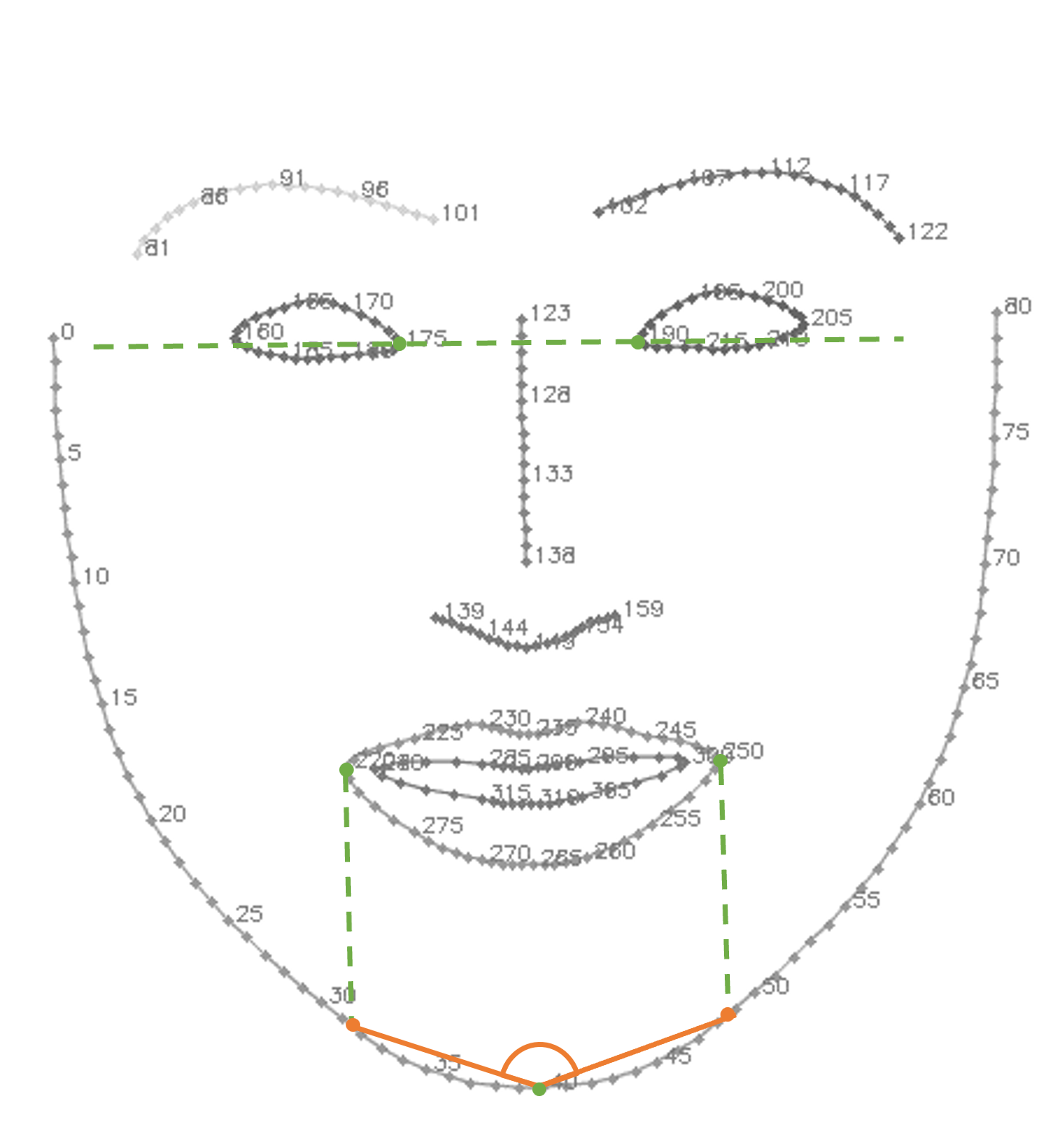}
}
\hfil
\subfigure[jaw area]{
    \centering
    \includegraphics[width=0.14\textwidth]{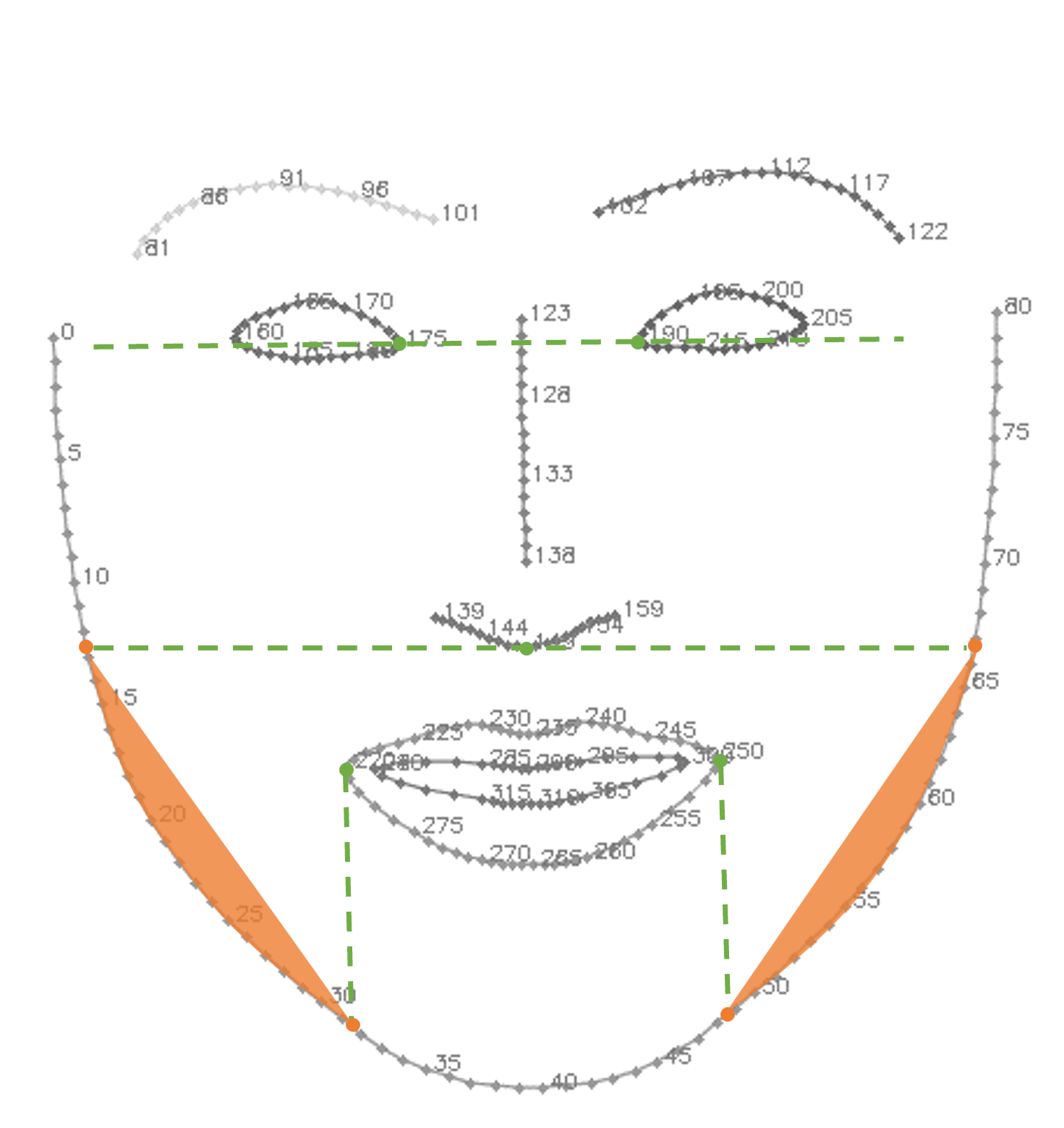}
}
\hfil
\subfigure[lip ratio]{
    \centering
    \includegraphics[width=0.14\textwidth]{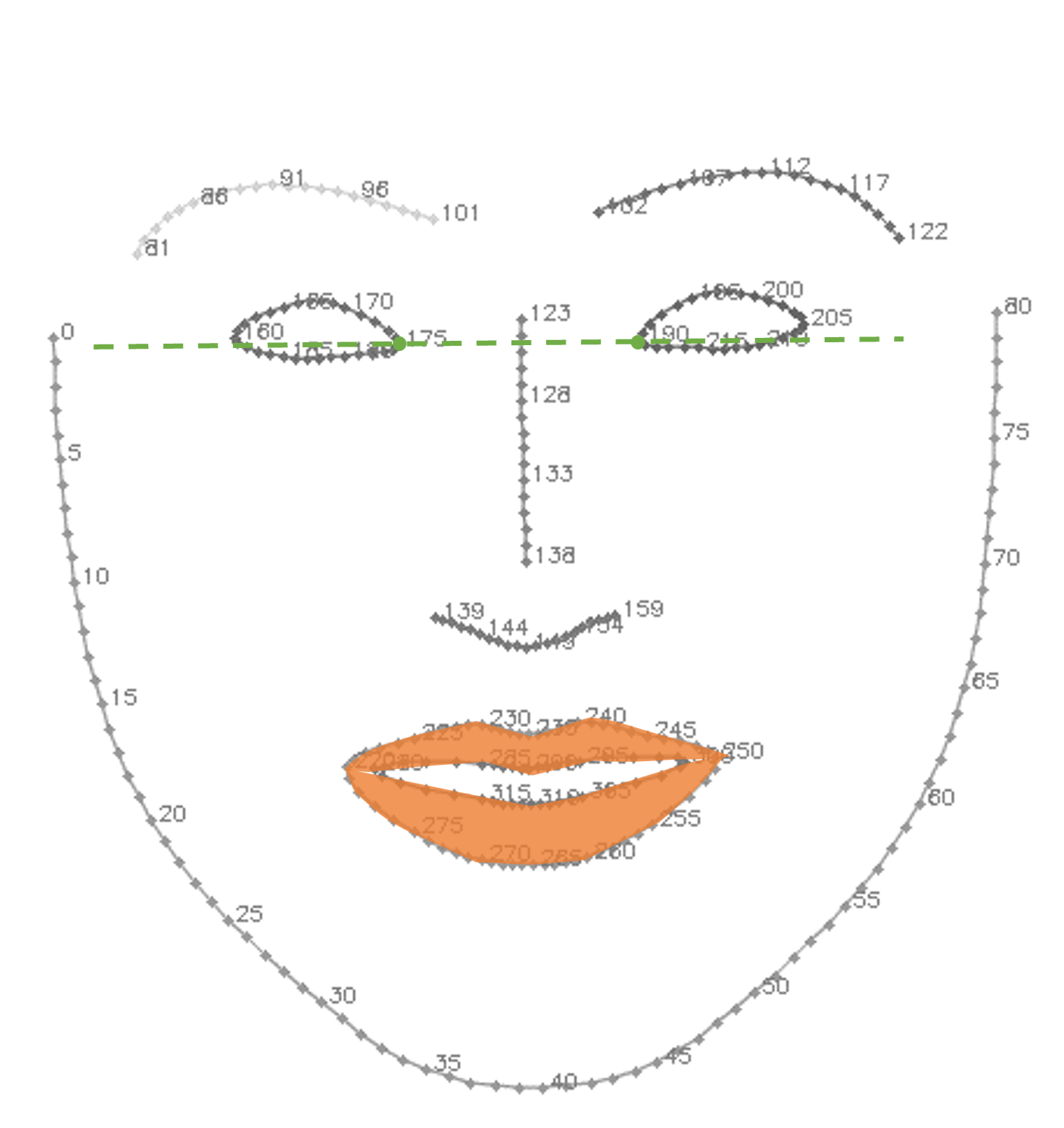}
}
\caption{The figure shows the definition of 4 facial morphometric measures based on semantically enriched landmarks. All measures have been normalized to a unit length which is the distance between the inner corner of the left and right eye.}
\label{figure:metrics_definition}
\end{figure}

Facial morphometric measures are widely used in cosmetic medicine to help with disease diagnosis and aesthetic analysis.
While the measures are usually calculated from landmarks, especially dense landmarks.
To illustrate the significance of enriching landmarks on downstream tasks, several facial morphometric measures, chin angle, jaw area, and the ratio of the upper and lower lip area, are designed based on existing facial landmarks in Fig~\ref{figure:metrics_definition}.
These facial morphometric measures are intuitive and discriminative on the face images of the 300W dataset. 




By the MAPE in Table~\ref{table:metrics_comparison}, the predicted enriched landmarks outperform sparse landmarks even in ground-truth landmarks, when applying them to the facial morphometric measures.
To conclude, dense landmarks are better than sparse landmarks in the application of facial morphometric measures.
Enriching landmarks is meaningful for some downstream tasks.

\begin{figure}[H]
	\centering
	\subfigbottomskip=-2pt
	\subfigcapskip=-1pt
	\subfigure[chin angle samples]{
	    \centering
		\includegraphics[width=0.98\linewidth]{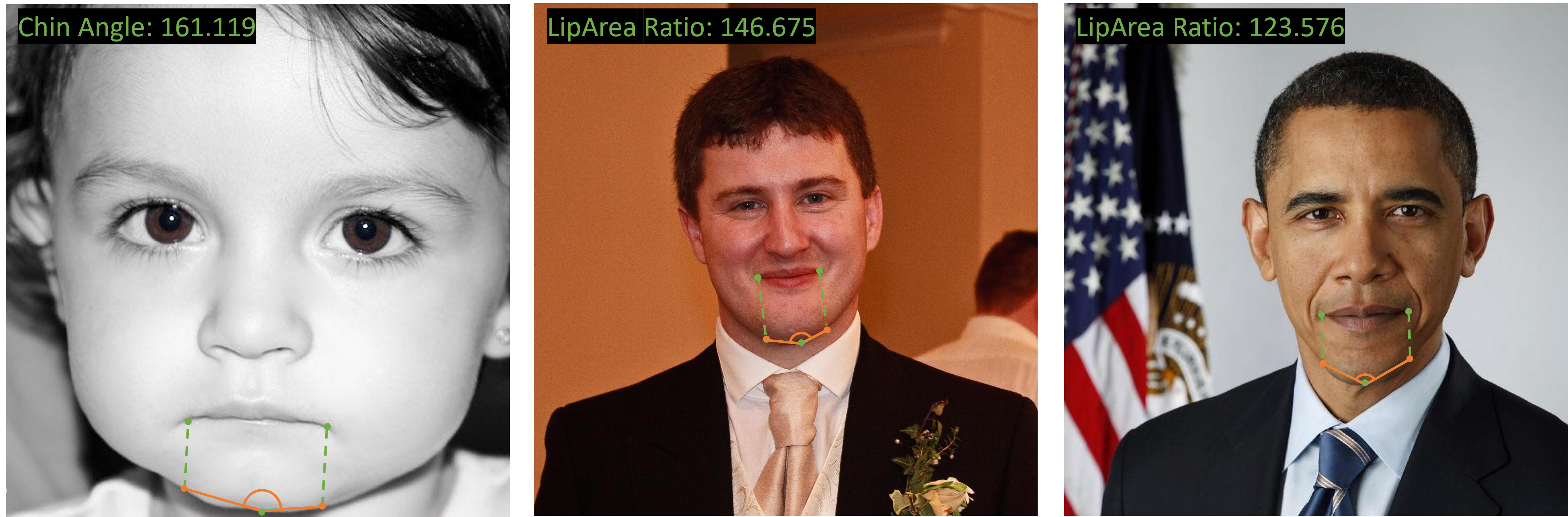}}

	\subfigure[left jaw area samples]{
	\centering
		\includegraphics[width=0.98\linewidth]{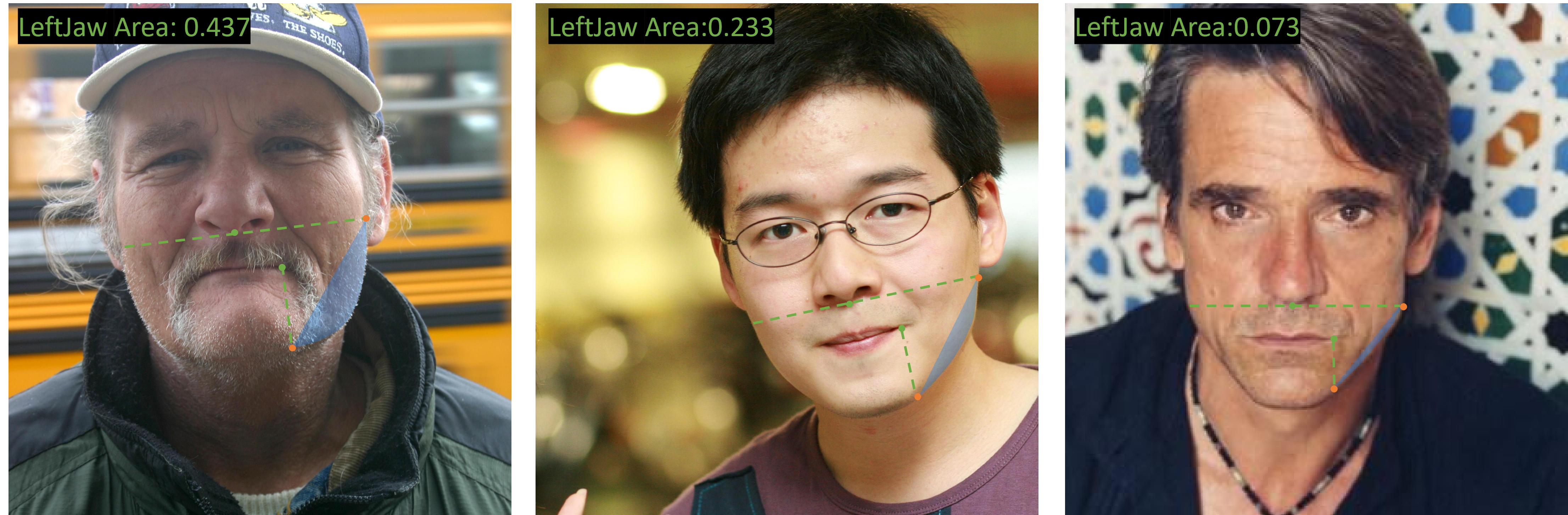}}

	\subfigure[right jaw area samples]{
	\centering
		\includegraphics[width=0.98\linewidth]{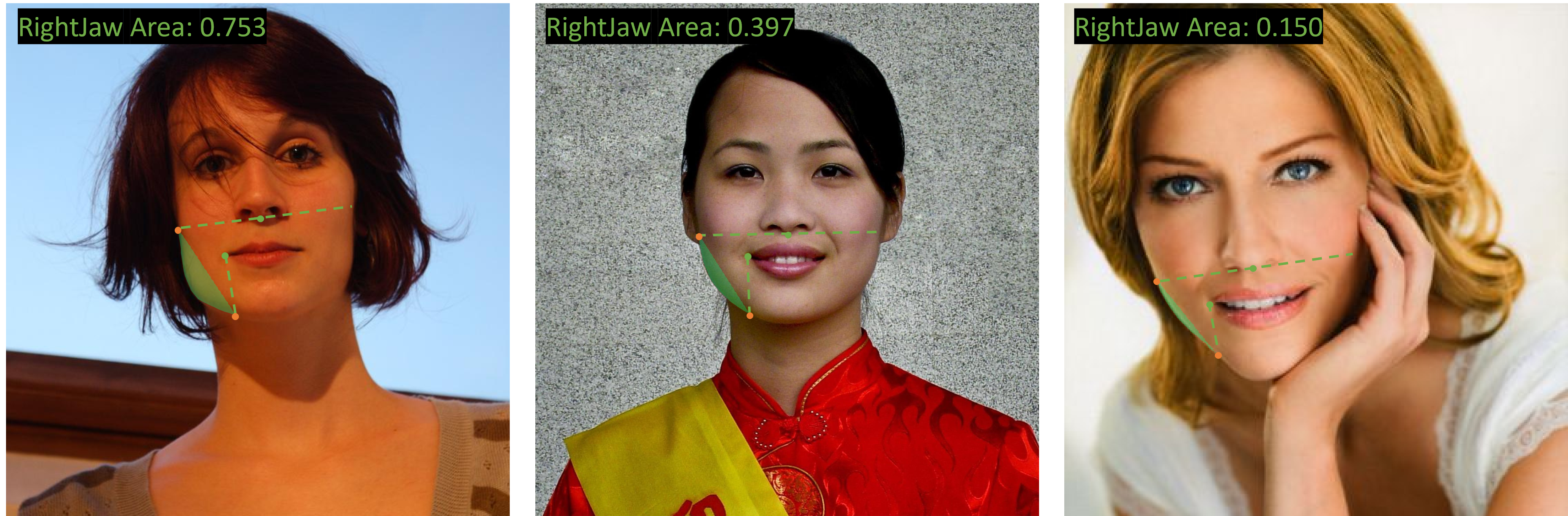}}

	\subfigure[lip area ratio samples]{
	\centering
		\includegraphics[width=0.98\linewidth]{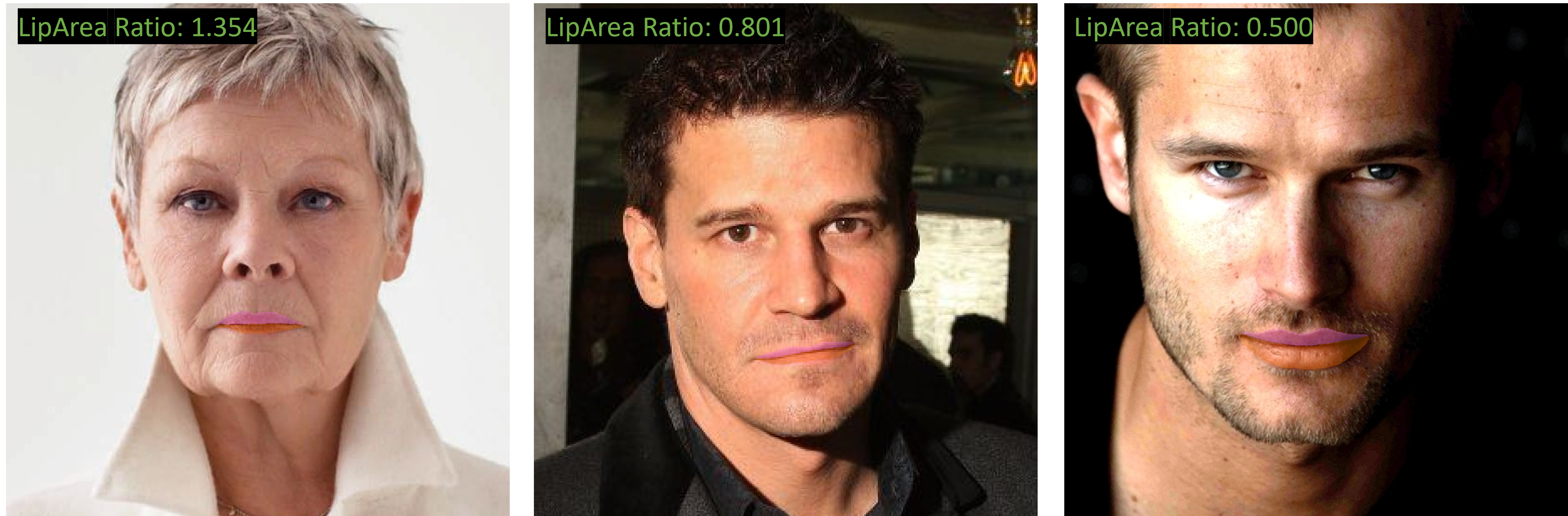}}
	\caption{Here are several examples of how well these measurement can be distinguished on 300W face images. The chin angle describes the sharpness of the chin, the jaw area presents the fatness of the jaw, and the lip area ratio indicates whether the upper lip is thinner or thicker than the lower lip. }
\end{figure}

\begin{table}[htbp]
\centering
\small
\resizebox{\columnwidth}{!}{
\begin{tabular}{llllll}
\hline
Method & Landmarks & \makecell{Chin \\ Angle} & \makecell{LeftJaw \\ Area} & \makecell{RightJaw \\ Area} & \makecell{LipArea \\ Ratio} \\
\hline
ADNet \cite{huang2021adnet} & Sparse & 1.414 & 5.950 & 6.223 & 5.146 \\
Ground-truth & Sparse & 1.317 & 5.355 & 5.483 & 5.514 \\
\textbf{ADNet-FE5 (Ours)} & \textbf{dense} & \textbf{0.369} & \textbf{3.189} & \textbf{3.091} & \textbf{4.373} \\
\hline
\end{tabular}
}
\caption{MAPE Comparison of different methods on 300W dataset. 4 facial morphometric measures are demonstrated in total.}
\label{table:metrics_comparison}
\end{table}


\section{Additional Details}

\subsection{Landmark Contour Definition}
\begin{figure}[h]
\small
\centering
\includegraphics[width=0.85\linewidth]{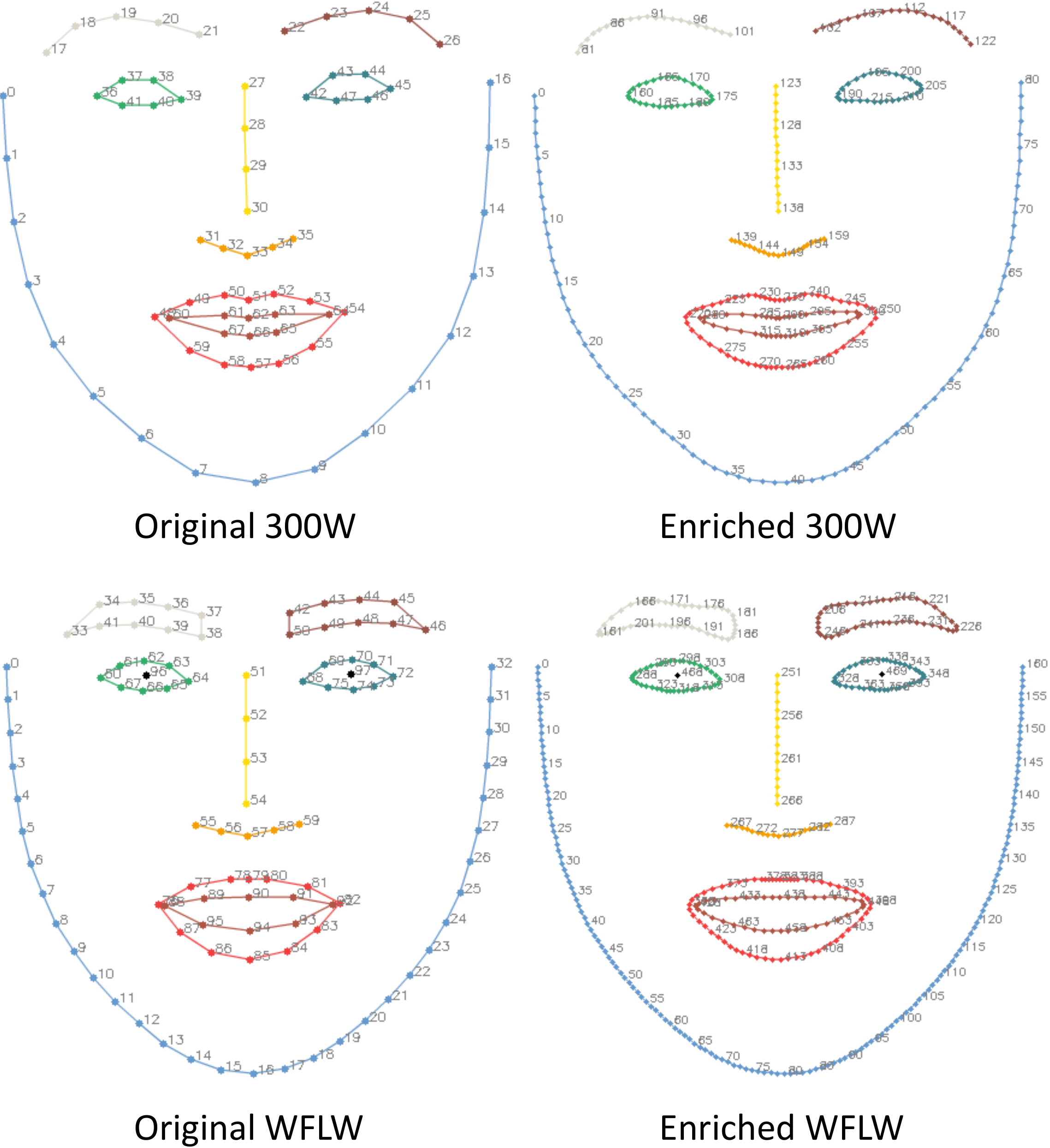}
\caption{Facial component definition in the 300W and WFLW datasets with/without FreeEnricher enrichment under the enriching density of 5. Different colors denote different components as defined in Table~\ref{table:edges_definition}.}
\label{figure:edges_definition}
\end{figure}

We divide the facial landmark points into several landmark contours according to their semantic relations, they are facial contour, left/right eyebrow, left/right eye, nose middle/bottom line, inner/outer lip, and left/right pupil.
For the 300W and WFLW datasets, there are 9 and 11 landmark groups, respectively, as shown in Table~\ref{table:edges_definition} and Fig.~\ref{figure:edges_definition}.
Except for the two isolated landmarks of pupils in WFLW, all the landmarks are located on an edge or a contour line.

\begin{table}[htbp]
\scriptsize
\begin{center}
\begin{tabular}{ccccc}
\hline
\multirow{2}*{\makecell{Facial \\ Component \\ Names}} & \multicolumn{2}{c}{300W} & \multicolumn{2}{c}{WFLW} \\
\cline{2-5}
 & \makecell{Original \\ Indices} & \makecell{Enriched \\ Indices} & \makecell{Original \\ Indices} & \makecell{Enriched \\ Indices} \\
\hline
\textbf{\textcolor[rgb]{0.4,0.6,0.8}{\makecell{Facial Contour}}} & 0-16 & 0-80 & 0-32 & 0-160 \\
\hline
\textbf{\textcolor[rgb]{0.85,0.843,0.8}{\makecell{Eyebrow Right}}} & 17-21 & 81-101 & 33-41 & 161-205 \\
\textbf{\textcolor[rgb]{0.6,0.33,0.29}{\makecell{Eyebrow Left}}} & 22-26 & 102-122 & 42-50 & 206-250 \\
\hline
\textbf{\textcolor[rgb]{0.99,0.863,0.067}{\makecell{Nose Middle Line}}} & 27-30 & 123-138 & 51-54 & 251-266 \\
\textbf{\textcolor[rgb]{0.95,0.623,0.0197}{\makecell{Nose Bottom Line}}} & 31-35 & 139-159 & 55-59 & 267-287 \\
\hline
\textbf{\textcolor[rgb]{0.2156,0.6784,0.4196}{Eye Right}} & 36-41 & 160-189 & 60-67 & 288-327 \\
\textbf{\textcolor[rgb]{0.243,0.5176,0.549}{Eye Left}} & 42-47 & 190-219 & 68-75 & 328-367 \\
\hline
\textbf{\textcolor[rgb]{0.937,0.25,0.25}{Lip Outer}} & 48-59 & 220-279 & 76-87 & 368-427 \\
\textbf{\textcolor[rgb]{0.694,0.3373,0.2745}{Lip Inner}} & 60-67 & 280-319 & 88-95 & 428-467\\
\hline
\textbf{\textcolor[rgb]{0.0,0.0,0.0}{Pupil Right}} & - & - & 96 & 468 \\
\textbf{\textcolor[rgb]{0.0,0.0,0.0}{Pupil Left}} & - & - & 97 & 469 \\
\hline
\textbf{Whole Face} & 0-67 & 0-319 & 0-97 & 0-469 \\
\hline
\end{tabular}
\end{center}
\caption{Landmark indices of different facial components in 300W and WFLW datasets before/after enrichment with the enriching density of 5.}
\label{table:edges_definition}
\end{table}

\subsection{Landmark Contour Fitting}
\begin{figure}[h]
\begin{center}
\includegraphics[width=3.3in]{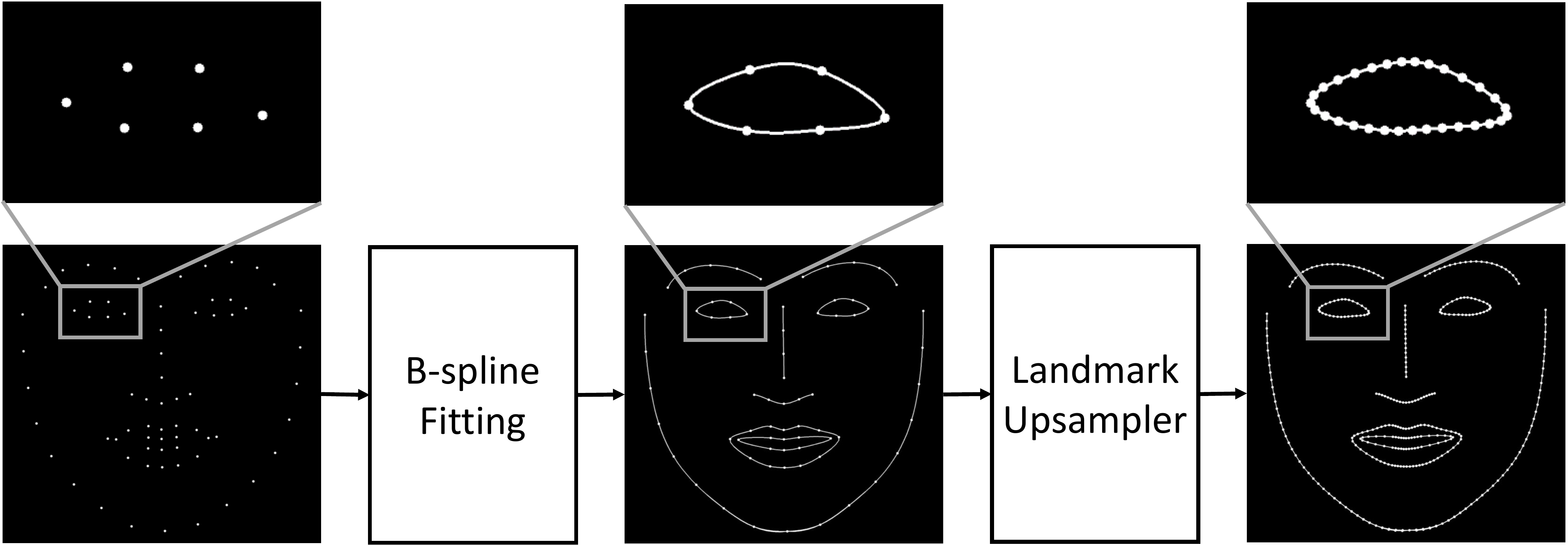}
\end{center}
   \caption{Illustration of landmark upsampling. Both straight line and b-spline could be applied to fit the continuous curve. Then the initialized enriched landmarks could be generated by average upsampling the fitted curve.
   }
\label{figure:landmark_contour_upsampling}
\end{figure}

To initialize the enriched facial landmarks, both straight line and b-spline~\cite{knott2000interpolating} could be adopted to roughly fit the facial contour in Fig.~\ref{figure:landmark_contour_upsampling}.
For instance of b-spline, it is a general form of bezier curve, which is a linear combination of control points and basic functions. B-spline has continuous derivatives in each split point but each segment has an independent shape. 
Given a set of ordered points, b-spline can be fitted as
\begin{equation}
C(u) = \sum\limits_{i=0}^{n+d} b_{d, i}(u) \cdot q_{i}
\label{equation:b-spline}
\end{equation}
where $q_i$ is the control point, $d$ is the order, $n$ is the number of knots, $b_{d,i}(\cdot)$ is the basis function, and $u$ is the input variable to control the position of a point in curve. 
For the given sparse landmarks belonging to one edge, each of them could be used as a knot to fit the curve.
Thus, b-spline is applied to each independent facial component having contour or line information, such as eye, mouth and face contour.

\begin{table*}[htbp]
\small
\begin{center}
\begin{tabular}{lccccccccccc}
\hline
\multicolumn{2}{c}{Facial Component Names} & \makecell{Facial \\ Contour} & \makecell{Eyebrow \\ Right} & \makecell{Eyebrow \\ Left} & \makecell{Nose \\ Middle \\ Line} & \makecell{Nose \\ Bottom \\ Line} & \makecell{Eye \\ Right} & \makecell{Eye \\ Left} & \makecell{Lip \\ Outer} & \makecell{Lip \\ Inner} & \makecell{Whole \\ Face} \\
\hline
\multicolumn{2}{c}{Avg. $S$} & 7.719 & 4.009 & 3.931 & 1.267 & 1.267 & 2.968 & 2.061 & 3.013 & 3.139 & 3.964 \\
\hline
\multirow{2}*{NME$_{point}$} & ADNet+Line5 & 5.617 & 3.240 & 3.125 & 1.990 & \textbf{2.538} & 1.639 & 1.601 & 2.080 & 4.014 & 3.208 \\
 & \textbf{ADNet-FE5} & \textbf{5.282} & \textbf{2.984} & \textbf{2.848} & \textbf{1.888} & 2.618 & \textbf{1.490} & \textbf{1.472} & \textbf{1.889} & \textbf{3.711} & \textbf{3.061} \\
\hline
\multirow{2}*{NME$_{edge}$} & ADNet+Line5 & 1.541 & 1.136 & 1.134 & 0.880 & 1.270 & 0.934 & 0.932 & 0.988 & 0.973 & 1.183 \\
 & \textbf{ADNet-FE5} & \textbf{1.374} & \textbf{1.058} & \textbf{1.060} & \textbf{0.873} & \textbf{1.208} & \textbf{0.666} & \textbf{0.636} & \textbf{0.827} & \textbf{0.757} & \textbf{0.976} \\
\hline
\end{tabular}
\end{center}
\caption{$\textnormal{NME}$ Evaluation of different facial components on \textbf{enriched} 300W testing set.}
\label{table:components}
\end{table*}

\begin{table*}[hbtp]
\small
\begin{center}
\begin{tabular}{lccccccc}
\hline
Method & Testset & \makecell{Pose \\ Subset} & \makecell{Expression \\ Subset} & \makecell{Illumination \\ Subset} & \makecell{Make-up \\ Subset} & \makecell{Occlusion \\ Subset} & \makecell{Blur \\ Subset} \\
\hline
LAB \cite{wu2018look} & 5.27 & 10.24 & 5.51 & 5.23 & 5.15 & 6.79 & 6.12 \\
HRNet \cite{wang2020deep} & 4.60 & 7.86 & 4.78 & 4.57 & 4.26 & 5.42 & 5.36 \\
ADNet \cite{huang2021adnet} & 4.14 & 6.96 & \textbf{4.38} & 4.09 & 4.05 & 5.06 & 4.79 \\
\textbf{ADNet-FE5 (Ours)} & \textbf{4.10} & \textbf{6.94} & \textbf{4.38} & \textbf{4.07} & \textbf{3.91} & \textbf{4.92} & \textbf{4.71} \\
\hline
\end{tabular}
\end{center}
\caption{$\textnormal{NME}$ comparison between our method and other state-of-the-arts on different scenario testsets of the \textbf{original} WFLW.}
\label{table:WFLW_origin}
\end{table*}

\subsection{Landmark Normal Calculation}
Based on the fitted curve, denoted by $C(u)$, the derivative $C'(u)$ of the curve can be calculated by following  \cite{prochazkova2005derivative}.
Hereon, $C'(u)$ is also a b-spline curve with the order of $d{-}1$ on the original knot vector and the new set of control points.
Then the normal direction of the enriched landmarks in the fitted b-spline could be derived by using $C'(u)$, which is a necessary dependence for the \emph{patch normalizing} operator in FreeEnricher.

\subsection{Regression Model Architecture}
\begin{figure}[b!]
\centering
\includegraphics[width=1\linewidth]{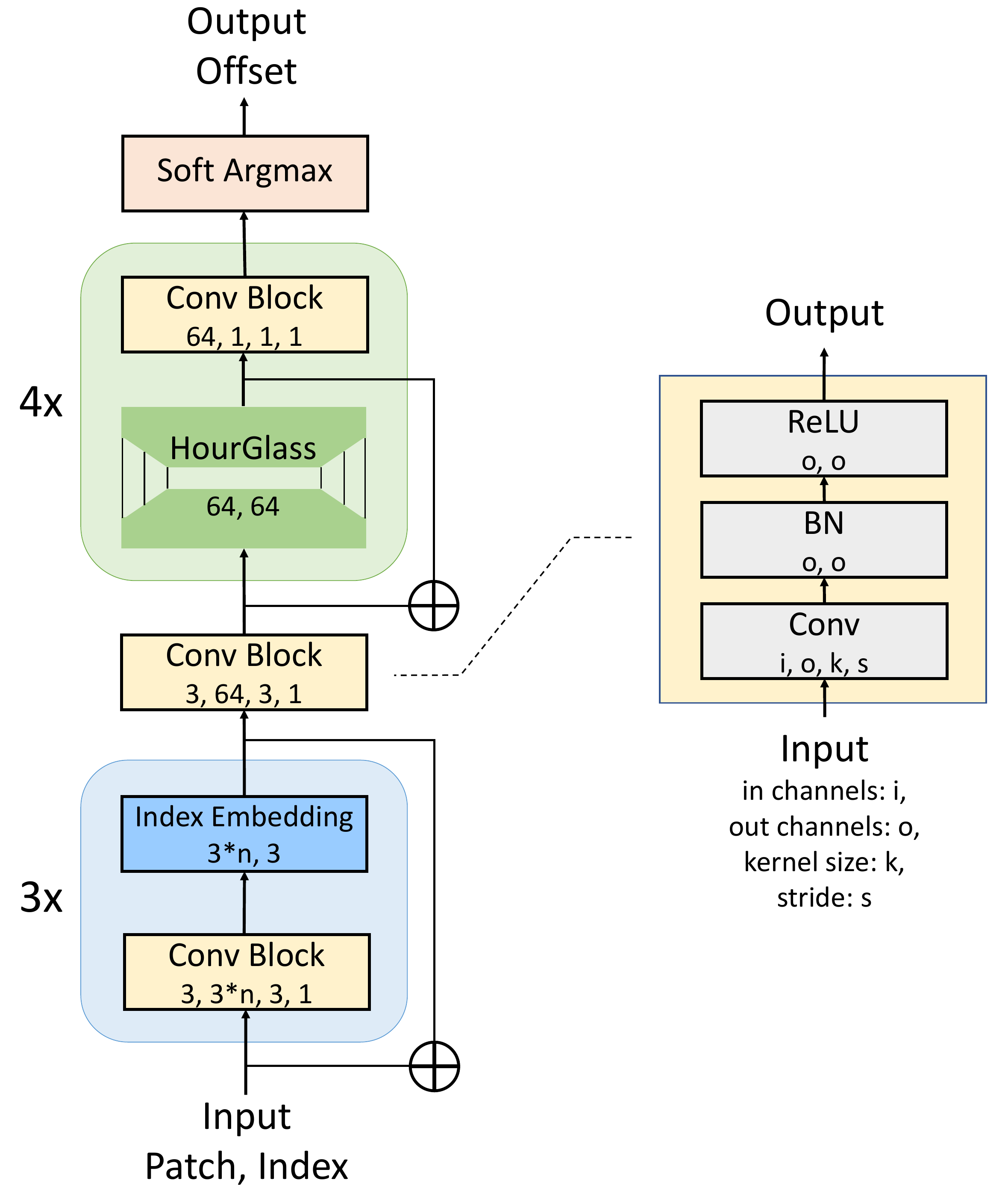}
\caption{Architecture of the offset regression model in FreeEnricher framework.
The left part demonstrates the detailed construction. The right part shows the $Conv\ Block$ used as a submodule. $n$ denotes the number of original landmarks in the dataset.
}
\label{figure:architecture_LRM}
\end{figure}

Fig.~\ref{figure:architecture_LRM} demonstrates the detailed architecture of the offset regression model. 
Except for the $Index\ Embedding$ layer that is devised in this paper, the other layers are all widely-used layers.
In the case of $Conv\ Block$ and $Conv\ Layer$, there are 4 hyperparameters which consist of the number of input channels, the number of output channels, the kernel size, and the stride respectively.
$Index\ Embedding$ layer, $HourGlass$ module, $BN$ (Batch normalization) layer and $ReLU$ layer have 2 hyperparameters which are the number of input and output channels.

\section{Additional Experiments}

\subsection{Evaluation on Different Components}
Apart from evaluating the performance of the whole facial landmarks, we also provide the individual results on different facial components defined in Table~\ref{table:edges_definition}.
As we only have the ground truth for the enriched 300W testing set, the experiments are conducted on it as tabulated in Table~\ref{table:components}.
Compared with baseline ADNet, there are two important observations. First, FreeEnricher has more improvement on NME$_{edge}$($\sim$20\%) than on NME$_{point}$($\sim$5\%), which indicates that the improvement along the normal direction is more significant. Second, FreeEnricher results in great accuracy improvement on facial components having obvious edge features, such as facial contours and eyes.

\subsection{Evaluation on Extreme Scenarios}
To verify the effectiveness of our method in extreme scenarios, we conduct experiments on subset of WFLW testset.
For the reason, $Quality\ Scoring$ would assign a low weight to the extreme cases, and $Landmark\ Refining$ would adjust their location carefully by the weight, the performance of those points are stable or even better, which is demonstrated in Table~\ref{table:WFLW_origin}.

\subsection{Comparison of Visualized Results}
\begin{figure*}[htbp]
\begin{center}
\includegraphics[width=6.6in]{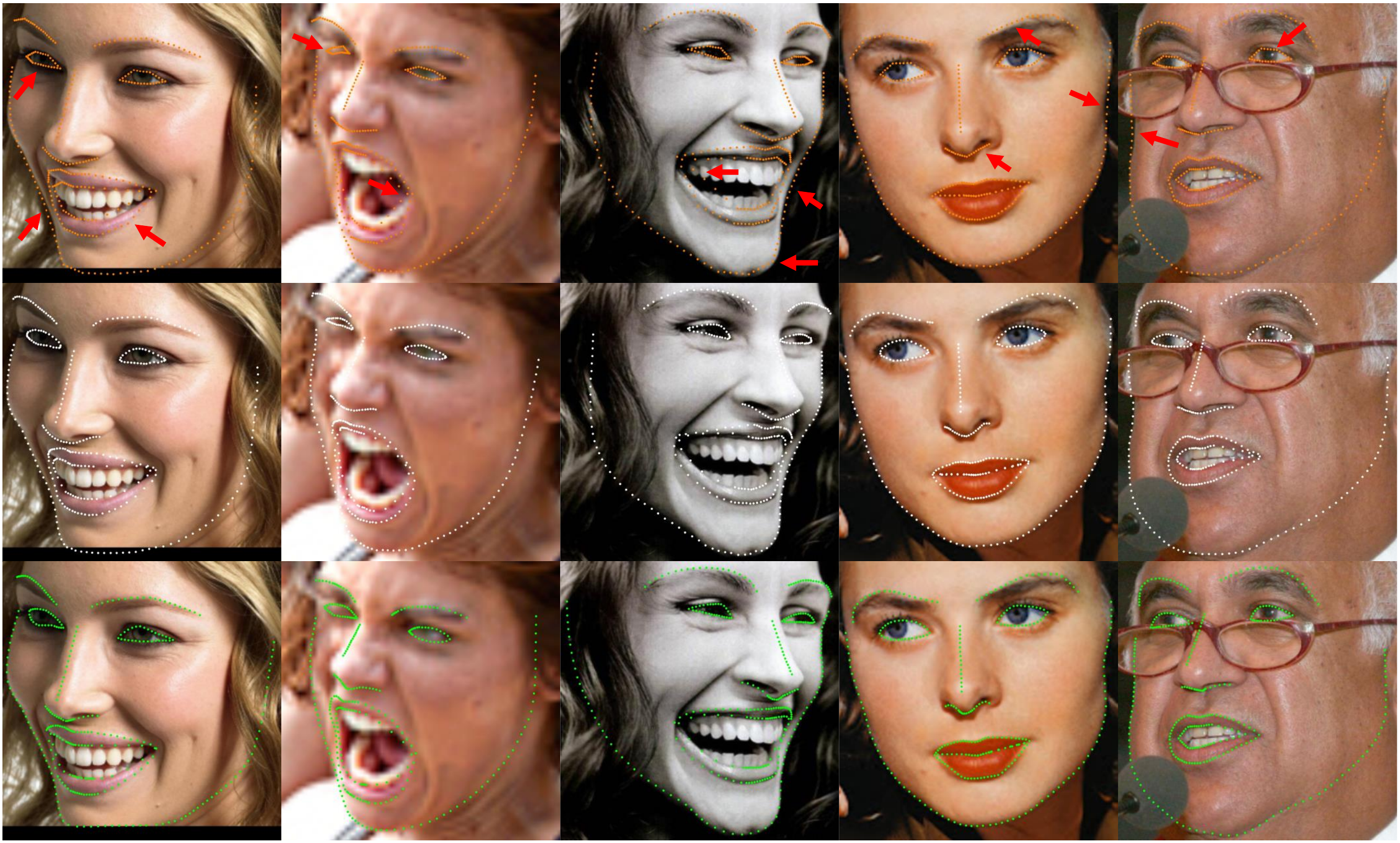}
\end{center}
\caption{Samples of visualized results in enriched 300W test dataset.
For each row, the first row means the results of baseline ADNet with linear interpolation, the second row shows the results of ADNet-FE5, and the last row indicates the ground truth labels made by manual annotation.
}
\label{figure:result_300W}
\end{figure*}

\begin{figure*}[hbtp]
\begin{center}
\includegraphics[width=6.6in]{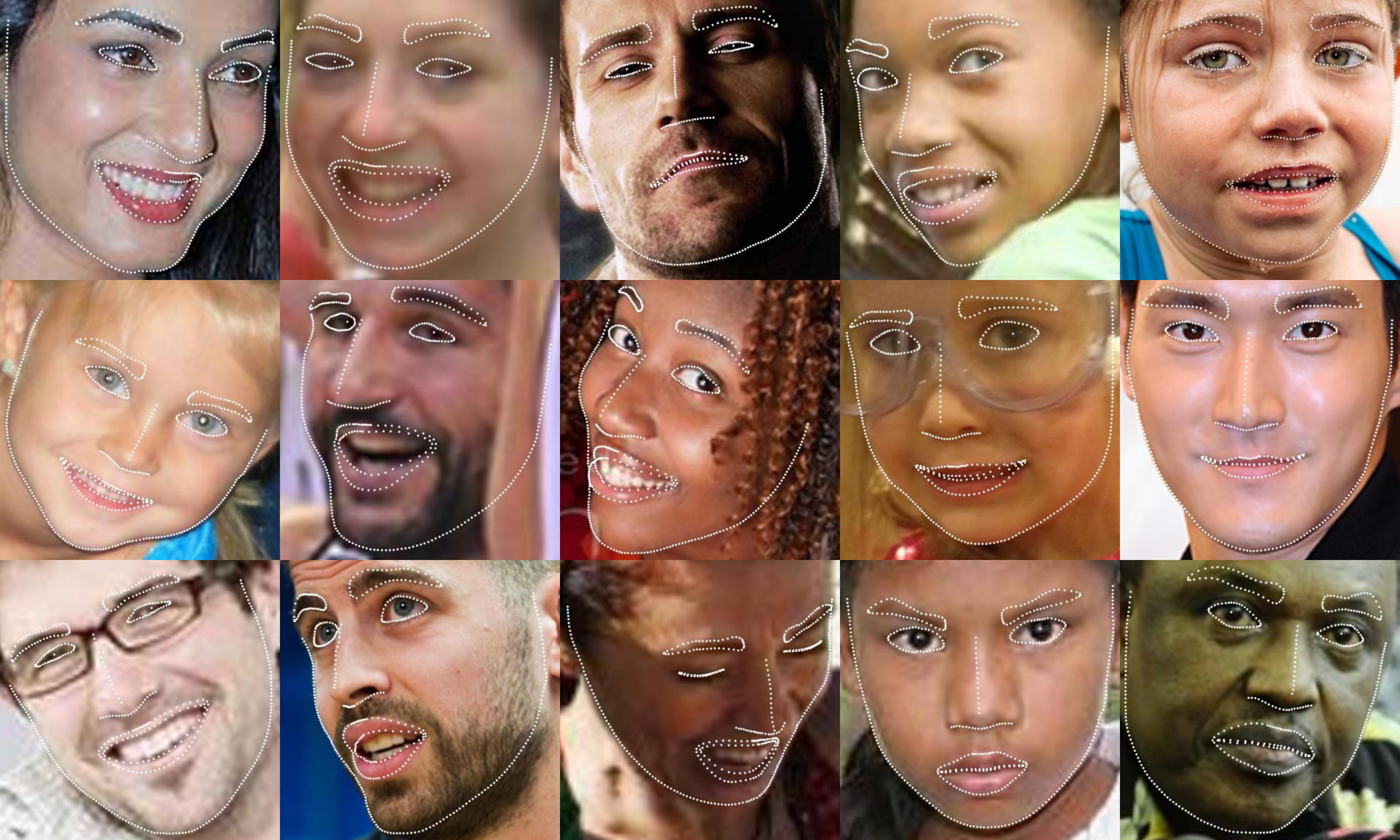}
\end{center}
\vspace{-10px}
\caption{Samples of visualized results detected by ADNet-FE5 in WFLW test dataset.
}
\label{figure:result_WFLW}
\end{figure*}

To show the extended qualitative results, we randomly select samples in 300W testing dataset in Fig.~\ref{figure:result_300W}.
We observe that ADNet-FE5 infers landmarks more tightly to the corresponding contours than the baseline method, and there is no regression in the occluded or blurry regions.
Several samples on WFLW are also illustrated in Fig.~\ref{figure:result_WFLW}, where only the inference results of ADNet-FE5 are shown since there is no ground truth of the enriched landmark.
It is recommended to zoom in the images for better discrimination.

\clearpage

\end{document}